# Meta-Learning and Knowledge Discovery based Physics-Informed Neural Network for Remaining Useful Life Prediction

Yu Wang, Shujie Liu*, Shuai Lv, Gengshuo Liu

*State Key Laboratory of High-performance Precision Manufacturing, Dalian University of Technology, Dalian 116024, China*

Abstract: Accurately predicting the Remaining Useful Life (RUL) of critical machinery is essential for enhancing the reliability and operational efficiency of modern manufacturing systems. However, existing machine learning-based methods often face two major challenges in practical applications: scarcity of target domain samples and a lack of explicit dynamic equation descriptions for the equipment degradation process, making it difficult for models to effectively incorporate physical laws for prediction. To address these challenges, this paper proposes a Meta-Learning and Knowledge Discovery based Physics-Informed Neural Network (MKDPINN) for RUL prediction of machinery. The method first maps high-dimensional, noisy sensor data to a smooth, low-dimensional hidden state space characterizing equipment degradation through a Hidden State Mapper (HSM). Subsequently, a Physics-Guided Regulator (PGR) adaptively learns the unknown non-linear partial differential equation (PDE) governing its evolution based on this hidden state, realizing the discovery of potential physical laws from the data. This discovered PDE is embedded as a physical constraint into the PINN framework, guiding RUL prediction through a regularized loss function, and deeply integrating data-driven and physics knowledge. This framework is further embedded into a first-order meta-learning paradigm, and by performing inner and outer loop optimization on multiple source domain meta-tasks, the gradient of the expected loss function of all meta-tasks is implicitly approximated, aligning the model parameters with the aggregated optimal parameters from each meta-task, requiring only a few target domain samples to quickly adjust the model and adapt to new RUL prediction tasks. Experimental validation using data from industrial slurry pumps operating in a large-scale processing system and the public C-MAPSS benchmark confirms the superior performance and generalization of the proposed MKDPINN compared to baseline models. **Source code of the proposed model is publicly available at** https://github.com/Sephiroth66616/MKDPINN

## 1. Introduction

Modern manufacturing systems, increasingly reliant on automation and complex equipment integration, depend heavily on the reliable operation of core components like rotating machinery. [1]. These devices operate under high loads and complex working conditions for extended periods, inevitably leading to various forms of degradation and failure. Within the context of interconnected manufacturing environments, unscheduled downtime of such critical machinery not only results in substantial economic losses due to production halts but can also cascade through the system, potentially triggering safety incidents. Therefore, accurately predicting the Remaining Useful Life (RUL) of rotating machinery is crucial for enabling data-driven maintenance strategies, ensuring the overall reliability and resilience of the manufacturing system, optimizing operational schedules, and reducing lifecycle costs [2, 3].

In recent years, with the rapid development of prognostics and health management (PHM)

technology, RUL prediction methods based on machine learning (ML) have gradually become a research hotspot and have demonstrated great application potential[4]. These methods do not rely on an in-depth understanding of the degradation mechanisms of the equipment, but instead learn degradation patterns directly from historical operating data [5] and predict the RUL of the equipment accordingly [6]. Compared to traditional methods based on physical models and statistics-based models, this approach exhibits strong nonlinear fitting capabilities, adaptability, and the potential for modeling complex mechanical systems [7]. For example: Shi et al.[8] proposed a lightweight dual-attention mechanism and long short-term memory network model (DA-LSTM), combining LSTM's ability to model sequential degradation features with the dual-attention mechanism's ability to aggregate features, achieving accurate RUL prediction; Zhong et al.[9] proposed a bearing RUL prediction model based on a multi-region hypergraph self-attention network (M-HGSAN), effectively extracting high-order correlations and key features from bearing monitoring data by constructing multi-region receptive fields and combining hypergraph neural networks and self-attention mechanisms; Wu et al.[10] proposed a dual-path prediction architecture (DP-TSDDC) based on time series decomposition and degradation correction, which achieves effective mining and fusion of fine-grained information in the data by decomposing monitoring data into trend and non-stationary components, and using temporal convolutional networks and Patch-integrated LSTM with a multi-head attention mechanism to process them separately. However, the application of ML-based methods in real-world industrial scenarios still faces many challenges.

 One of the challenges is the scarcity of target domain samples[11]. In real-world industrial environments, equipment typically operates under complex and variable working conditions, and for safety and economic reasons, obtaining large amounts of high-quality degradation data (especially full life-cycle data from healthy state to complete failure) is often extremely challenging [12, 13]. Although sufficient training data may exist in other related tasks or equipment (i.e., the source domain) [14], directly applying models trained on the source domain to the target domain (e.g., new model equipment, new working conditions, etc.) often leads to domain shift problems [15], resulting in decreased prediction performance and limiting the reliability of the model in practical applications [16]. To address this challenge, researchers have attempted to introduce Transfer Learning (TL) [17] and Meta-Learning [18] technologies. TL uses Domain Adaptation (DA) [19] methods to transfer a pre-trained model from a source domain (typically with sufficient data) to a target domain, to achieve cross-domain reuse of knowledge [20, 21]. For instance, Xiang et al. [22] proposed a micro-transfer learning mechanism for multiple differentiated distributions, using multi-unit LSTMs to obtain multiple differentiated distributions of monitoring data at different time points, and using a domain adversarial mechanism to achieve knowledge transfer at the unit level, solving the unreasonable assumption of using a single distribution to represent time-varying life cycle signals in traditional transfer learning; Li et al. [23] proposed a segmented adaptive RUL prediction method based on multi-feature space cross-adaptation transfer learning, improving the RUL prediction accuracy under unseen degradation data; Lyu et al.[24] proposed a TL method based on deep degradation feature adaptive alignment, which solves the problem of traditional domain adaptation methods ignoring the influence of local features and improves the performance of RUL prediction under different working conditions. However, the effectiveness of TL is highly dependent on the selection of the transfer strategy and the similarity between the source and target domains [25]. When the inter-domain differences are significant, simple transfer may lead to Negative Transfer, which can reduce the performance of the model in the target domain. In addition, traditional TL methods usually require redesigning the transfer strategy for each new target domain, lacking flexibility and generality. In contrast, Meta-Learning, also known as "Learning to Learn," provides a more flexible solution [26]. By learning a general initialization model or learning strategy on multiple related tasks, the model/strategy can quickly adapt to new tasks (target domain tasks) with only a few samples [27]. This method can not only effectively utilize the knowledge of the source domain, but also achieve rapid learning and generalization in the case of scarce target domain samples [28]. Currently, algorithms based on Model-Agnostic Meta-Learning (MAML) [29] have shown promising application prospects in few-shot RUL prediction [30, 31]. For example, Ding et al. [32] explored the application of prediction methods based on graph neural networks (GNNs) under partial monitoring data, established a graph-structured few-shot prediction framework, and on this basis proposed a graph prediction algorithm based on embedded meta-learning training models, realizing the processing of spatio-temporal graph data and cross-domain generalization ability in a few-shot training mode; Rai et al.[33] proposed a multi-stage, feature-adaptive meta-learning model and combined bidirectional long short-term memory networks (BiLSTM) and variational autoencoders (VAE) to capture statistical changes and time dependencies, achieving accurate prediction of RUL; Chang et al. [34] proposed a Bayesian Model-Agnostic Meta-Learning with Prediction Uncertainty Calibration (BMLPUC) few-shot learning method, which achieves accurate quantification and calibration of prediction uncertainty in data-scarce scenarios by introducing an uncertainty calibration term in the objective function of the model training. However, MAML-based meta-learning RUL prediction algorithms involve the computation of second-order derivatives during training, which brings huge computational burden and optimization instability [35], especially when dealing with high-dimensional data and complex models. Therefore, there is an urgent need to develop more concise and efficient meta-learning algorithms to address the challenges of rotating machinery RUL prediction in target domain sample scarcity scenarios.

Another major challenge lies in the lack of physics-informed information during the model training process [36]. Most existing methods mainly rely on statistical patterns in the data, while ignoring the rich physical knowledge contained in the equipment operation process [37, 38]. Some studies have attempted to incorporate physics-informed information into RUL prediction models [39, 40]. For example, Yang et al.[41] proposed a physics-informed driven bearing RUL prediction method, which realizes the effective extraction and fusion of sequential periodic features and medium-to-long-term trend features by constructing a physics-informed driven dynamic adaptive inverse discrete Fourier transform (IDFT) frequency domain module and a residual self-attention multi-state gated control unit; Wang et al.[42] proposed a general degradation physics-informed driven self-data-driven mechanical prediction method called Phyformer, which realizes the complementary advantages of physical models and data-driven models by combining a deep learning model based on autocorrelation and Transformer architecture with multiple local physical models constructed within a sliding time window; de Beaulieu et al.[43] incorporated prior system knowledge and failure physics into training data through a hybrid data augmentation procedure and developed an unsupervised health indicator (HI) extraction method, which solves the dependence of current prediction methods on labeled target data; Xiong et al.[44] proposed a hybrid framework combining a controllable physics-informed driven data generation method and a deep learning prediction model, generating physically interpretable and diverse synthetic degradation trajectories, which solves the problem that deep learning methods are limited in performance or cannot be applied in practical applications due to limited representative failure time (TTF) trajectory data.

However, these methods mostly remain at a shallow level of integration between physics-informed information and ML, that is, simple concatenation or weighting is mainly performed at the data level or model structure level. The reason is that for the complex degradation process of equipment, the derivation of first principles (such as establishing accurate partial differential equations) is difficult, and even in most scenarios, it is impossible to derive effective dynamic equations between sensor monitoring data and equipment degradation states. The lack of utilization of potential physical processes can lead to the model over-relying on observation data [44, 45], and the prediction robustness is insufficient when facing noise interference, changing working conditions, and other situations [46, 47].

Based on the above discussion, to solve the problems of insufficient target domain samples and the lack of physics-informed information during the training process in RUL prediction tasks in real industrial scenarios, we propose a Meta-Learning and Knowledge Discovery based Physics-Informed Neural Network (MKDPINN). Based on the idea of Deep Hidden Physics Models (DeepHPM) [48, 49], we construct a hierarchical learning framework composed of a Hidden State Mapper (HSM) and a Physics-Guided Regulator (PGR), where the HSM can map high-dimensional, noisy, multi-source sensor time series to a continuous latent state space, achieving effective representation of the equipment degradation state and avoiding the instability of traditional numerical differentiation methods in noisy environments. Based on the hidden state output by the HSM, the PGR adaptively learns the non-linear partial differential equation (PDE) followed by the evolution of the hidden state, which serves as the physical law constraint of equipment degradation. The PGR is embedded as a PDE into the PINN framework to guide the training process of the proposed method, thereby realizing the deep integration of data-driven and physics knowledge discovery. On this basis, aiming at the target domain data scarcity scenario, the proposed method is further extended to a first-order meta-learning framework [50]. By performing multi-step inner-loop gradient updates and outer-loop meta-parameter updates on multiple meta-tasks (each meta-task has only a few samples), the gradient of the expected loss function of all meta-tasks is implicitly approximated, so that the model parameters are aligned with the aggregated optimal parameters from each meta-task, enabling the proposed method to quickly adjust parameters with only a small amount of data from new equipment/working conditions to adapt to new RUL prediction tasks, while avoiding the high-order derivative calculations in MAML.

The main contributions of this paper are as follows:

1. We propose a hierarchical learning framework consisting of an HSM and a PGR, where the HSM is responsible for extracting smooth hidden states from raw sensor data, and the PGR can adaptively learn the non-linear PDE dynamics equation followed by the latent states.

2. We embed the PGR as a physical constraint into the PINN framework, guiding the training of the neural network through a regularized loss function, realizing the collaborative optimization of data-driven and physics knowledge, and improving the generalization ability of the model and its compliance with physical rules.

3. The model is further extended into a meta-learning framework, enabling it to quickly adjust model parameters with only a small amount of data from new equipment or new working conditions, achieving accurate prediction of new tasks, and avoiding the calculation of second-order derivatives based on MAML.

4. Comprehensive experimental validation is performed on real industrial scenarios and the publicly available CMAPSS dataset. The results show that MKDPINN exhibits superior performance in predicting RUL of the same equipment in different life cycles and across different equipment in the scenario of scarce target domain samples, verifying the effectiveness and robustness of the proposed method.

The subsequent organization of this paper is as follows: Section 2 reviews the relevant background; Section 3 elaborates on the theoretical framework and implementation details of the proposed MKDPINN method; Section 4 verifies and analyzes the performance of MKDPINN in different scenarios through experiments; Section 5 summarizes the full paper and looks forward to future research directions.

## 2. Problem Formulation

### 2.1 Physics-Informed Neural Networks

PINNs are a deep learning framework that embeds prior physical knowledge into neural networks, providing a promising approach to address the problem of missing physics-informed information during model training[38]. The core idea of PINNs is to incorporate PDEs that govern the behavior of control systems into the loss function of the neural network as regularization terms or soft constraints, thereby guiding the network to learn solutions that satisfy physical laws.

Assume that the degradation process of rotating machinery equipment can be described by the following partial differential equation:

$$\frac{\partial u(\boldsymbol{h},t)}{\partial t} = \mathcal{N}[u(\boldsymbol{h},t)], \quad (\boldsymbol{h},t) \in \Omega \times [0,T] \tag{1}$$

Where $u(\boldsymbol{h},t) \in R^d$ represents the state vector related to equipment degradation, which is a function of the hidden state $\boldsymbol{h} \in \Omega \subset R^{d_h}$ and time $t \in [0,T]$, where $d$ represents the dimension of the state vector. Here, the hidden state $\boldsymbol{h}$ represents those unobservable, but decisive, intrinsic state variables that drive the overall degradation behavior of the equipment (related to factors such as changes in the microstructure of the material, the degree of damage accumulation, and the evolution of internal defects). $\mathcal{N}[\cdot]$ represents a non-linear differential operator, which describes the physical process controlling equipment degradation (e.g., Navier-Stokes equations, heat conduction equations, etc.). $\Omega$ represents the hidden state space.

PINNs approximate the solution $u(\boldsymbol{h},t)$ of the PDE by constructing a neural network $u_{\boldsymbol{\theta}}(h,t)$, where $\boldsymbol{\theta}$ represents the parameters of the neural network (weights and biases). The input of the network is the hidden state $\boldsymbol{h}$ and time $t$, and the output is the equipment state vector $u_{\boldsymbol{\theta}}(\boldsymbol{h},t)$. The loss function of PINNs is usually a weighted combination of a data loss term and a physics loss term:

$$\mathcal{L}(\boldsymbol{\theta}) = w_d \mathcal{L}_{data}(\boldsymbol{\theta}) + w_p \mathcal{L}_{physics}(\boldsymbol{\theta}) \tag{2}$$

Where, $w_d$ and $w_p$ are the weighting coefficients of the data loss and the physics loss, respectively.

The data loss term $\mathcal{L}_{data}(\theta)$ measures the difference between the neural network prediction value and the observed data. For prediction-related tasks involving the equipment degradation process, the Mean Squared Error (MSE) is usually used as the loss function:

$$\mathcal{L}_{data}(\boldsymbol{\theta}) = \frac{1}{N}\sum_{i=1}^{N}||u_{\boldsymbol{\theta}}(\boldsymbol{h}_i,t_i) - u_i||_2^2 \tag{3}$$

Where, $(\boldsymbol{h}_i,t_i)$ represents the hidden state and time point corresponding to the observed data, $u_i$ represents the degradation state of the equipment at that space-time point, $N$ is the number of observed data, and $||\cdot||_2$ represents the L2 norm of the vector.

The physics loss term $\mathcal{L}_{physics}(\boldsymbol{\theta})$ measures the deviation between the neural network prediction value and the PDE constraint. Using Automatic Differentiation technology, the partial derivatives of the neural network output $u_{\boldsymbol{\theta}}(\boldsymbol{h}, t)$ with respect to the input $\boldsymbol{h}$ and $t$ can be calculated, thereby constructing the PDE residual:

$$r(\boldsymbol{h}, t; \boldsymbol{\theta}) = \frac{\partial u_{\boldsymbol{\theta}}(\boldsymbol{h}, t)}{\partial t} - \mathcal{N}[u_{\boldsymbol{\theta}}(\boldsymbol{h}, t)] \tag{4}$$

The physics loss is usually defined as the L2 norm of the PDE residual:

$$\mathcal{L}_{physics}(\boldsymbol{\theta}) = \frac{1}{M} \sum_{j=1}^{M} \left|\left| r(\boldsymbol{h}_j, t_j; \boldsymbol{\theta}) \right|\right|_2^2 \tag{5}$$

Where, $(\boldsymbol{h}_j, t_j)$ is a set of collocation points within the hidden state space and time domain $\Omega \times [0, T]$, and M is the number of collocation points.

By minimizing the total loss function $\mathcal{L}(\boldsymbol{\theta})$, PINNs can balance between data-driven and physical constraints, thereby learning a solution that both conforms to the observed data and satisfies physical laws. This optimization process is usually implemented using gradient-based optimization algorithms (such as Adam, L-BFGS, etc.).

2.2 Meta-learning

In the RUL prediction task of rotating machinery equipment, traditional Machine Learning methods usually assume that the training data and the test data come from the same distribution. However, in real industrial scenarios, this assumption is often difficult to satisfy due to the diversity of equipment operating conditions, the complexity of degradation patterns, and the limitations of data acquisition. Degradation trajectories of different equipment or the same equipment under different operating conditions may have significant differences, leading to a decline in the generalization ability of the model under new, unseen working conditions.

Meta-learning, also known as "Learning to Learn," improves the model's ability to quickly adapt to new tasks by learning knowledge across multiple tasks. In the context of rotating machinery RUL prediction, RUL prediction under different equipment or different working conditions can be regarded as different tasks.

Assume that there is a set of RUL prediction tasks $\{\mathcal{T}_p\}_{p=1}^{P}$, where each task corresponds to different equipment or working conditions. Each task $\mathcal{T}_p$ has an associated dataset $\mathcal{D}_p = \{(\boldsymbol{x}_i^{(p)}, y_i^{(p)})\}_{i=1}^{N_p}$, where $\boldsymbol{x}_i^{(p)}$ represents the observed data of the $i$-th sample in the $p$-th task (e.g., vibration signals, temperature, pressure, etc. collected by the sensor), $y_i^{(p)}$ is the corresponding RUL ground truth, and $N_p$ is the number of samples in the $p$-th task. Due to differences in equipment or working conditions, the data distribution $P_p(x, y)$ of different tasks may be different.

The goal of meta-learning is to learn a meta-model $M_{\boldsymbol{\Phi}}$ from this set of RUL prediction tasks $\{\mathcal{T}_p\}_{p=1}^{P}$, where the meta-model is defined by parameters $\boldsymbol{\Phi}$. $M_{\boldsymbol{\Phi}}$ should be able to quickly adapt and make accurate RUL predictions on a new task $\mathcal{T}_{new}$ (drawn from a possibly different, held-out, test task distribution, $P_{test}(\mathcal{T})$), using only a small amount of data from that new task.

More formally, the meta-learning goal is to find the optimal meta-model parameters $\boldsymbol{\Phi}^*$, such that:

$$\boldsymbol{\Phi}^* = \arg\min_{\boldsymbol{\Phi}} E_{\mathcal{T}_p \sim P_{train}(\mathcal{T})} \left[ \mathcal{L}\left( f_{\boldsymbol{\theta}_p}; \mathcal{D}_p^{supp} \right) \right] \tag{6}$$

Where $P_{train}(\mathcal{T})$ represents the distribution of training tasks; $\mathcal{T}_p \sim P_{train}(\mathcal{T})$ represents sampling a task from the training task distribution; $f_{\theta_p}$ is the model adapted to task $\mathcal{T}_p$, with parameters $\theta_p$. These parameters are obtained by adapting the meta-model $M_\Phi$ using a small amount of support data $\mathcal{D}_p^{supp}$ from task $\mathcal{T}_p$:

$$\theta_p = G(M_\Phi, \mathcal{D}_p^{supp}) \tag{7}$$

Where $G$ represents the rapid adaptation process (e.g., a few steps of gradient descent). $\mathcal{D}_p^{supp} = \{(x_i^{(p)}, y_i^{(p)})\}_{i=1}^{N_p^{supp}}$ is a small subset of $\mathcal{D}_p$.

## 2.3 Few-shot learning

Few-shot Learning (FSL) refers to a learning paradigm in which the model can quickly adapt to new tasks and make accurate predictions with only a very small number of labeled samples. In the rotating machinery RUL prediction task, FSL can be formally defined as follows:

Given a new RUL prediction task $\mathcal{T}_{new}$, its corresponding support set $\mathcal{D}_{new}^{supp} = \{(x_i^{new}, y_i^{new})\}_{i=1}^{K}$ contains only $K$ samples, where $K$ is very small (usually $K$ is between 1-15). $x_i^{new}$ represents the observed data of the $i$-th sample in the new task, and $y_i^{new}$ is the corresponding RUL ground truth. This setting is called $K$-shot learning.

In this paper, the goal of FSL is to use the meta-model $M_\phi$ obtained based on meta-learning to quickly adapt to the new task $\mathcal{T}_{new}$, and obtain a parameterized model $f_{\theta_{new}}$ for this task:

$$\theta_{new} = g(M_\phi, \mathcal{D}_{new}^{supp}) \tag{8}$$

This model $f_{\theta_{new}}$ should be able to obtain a low prediction error on the query set $\mathcal{D}_{new}^{query} = \{(x_j^{new}, y_j^{(new)})\}_{j=1}^{M}$ of the new task:

$$\mathcal{L}(f_{\theta_{new}}, \mathcal{D}_{new}^{query}) = \frac{1}{M}\sum_{j=1}^{M} \mathcal{L}\left(f_{\theta_{new}}(x_j^{(new)}), y_j^{(new)}\right) \tag{9}$$

In practical applications, FSL can effectively solve the problem of insufficient samples for RUL

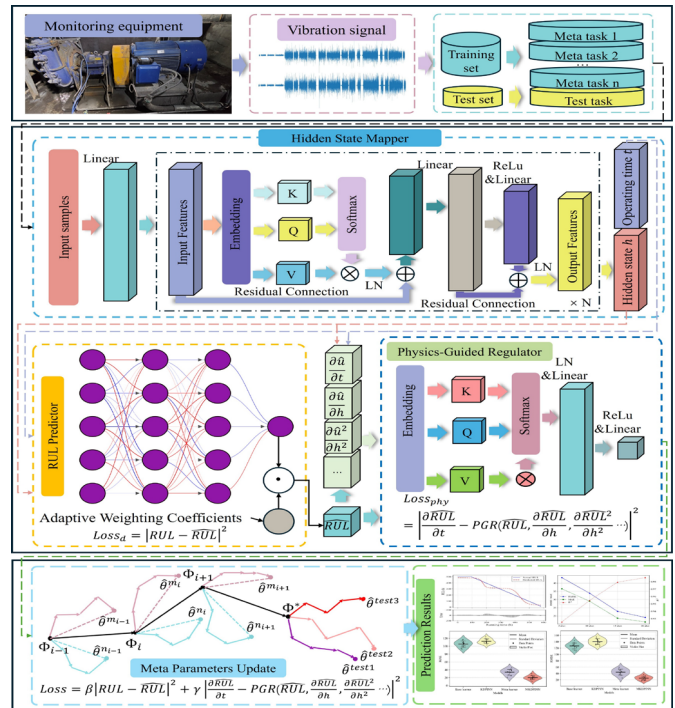

Fig. 1. The Overall Process of the Proposed MKDPINN

prediction under new equipment or new working conditions.

## 3. Proposed method

### 3.1 Overall framework of the proposed method

The overall workflow of the proposed MKDPINN is shown in Figure 1. First, the monitoring data during the equipment operation process is preprocessed, and a set of meta-tasks is constructed: the complete life-cycle data of each equipment is divided into multiple subsequences, and each subsequence constitutes an independent meta-task. Subsequently, the data flows through the three core modules of MKDPINN in sequence: the HSM, a self-attention-based neural network, maps the high-dimensional, noisy raw sensor data to a continuous hidden state space, extracting hidden state vector sequences that represent the equipment degradation dynamics; the PGR takes the hidden state sequence output by the HSM as input, and approximates the PDE followed by the hidden state evolution through a neural network; the RUL predictor combines the hidden state and running time information to predict the RUL of the equipment. The PDE learned by the PGR serves as a physical constraint, and together with the output of the RUL predictor, constitutes a loss function that includes a data-driven loss and a physical residual loss, thereby realizing the deep integration of the data-driven model and physics knowledge. The entire framework adopts a meta-learning strategy based on first-order gradient optimization for meta-level parameter updates: in the meta-training stage, multiple source domain tasks are sampled from the meta-task set for iterative updating, optimizing the global initial parameters; in the meta-testing stage, a small amount of target domain data is used for rapid adaptation, which can achieve rapid and accurate prediction of RUL under new equipment or new working conditions.

### 3.2 Knowledge Discovery based Physics-Informed Neural Network Framework

As shown in Figure 2, our proposed knowledge discovery physics-informed neural network framework consists of three key components: HSM, PGR, and RUL predictor, to achieve the fusion of data-driven and physical law constraints, and as a meta-learner in the subsequent meta-learning paradigm, it has the ability to quickly adapt to new RUL prediction tasks.

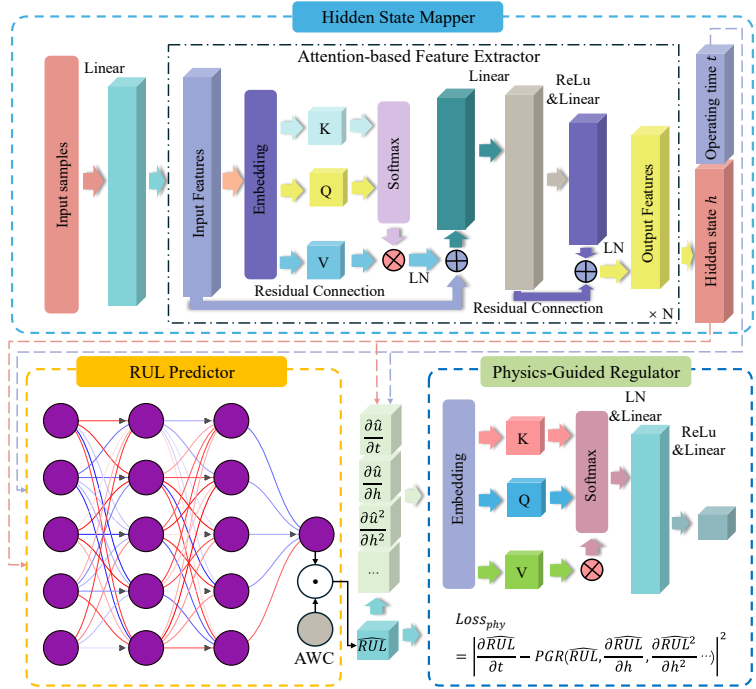

Fig. 2. Structure of the Knowledge Discovery based Physics-Informed Neural Network Framework

*3.2.1 Hidden State Mapper (HSM)*

In real industrial environments, the monitoring data of rotating machinery is usually high-dimensional, multi-source, and contains noise. Directly using raw sensor data for RUL prediction and physical law discovery faces many challenges, including the ill-posedness of numerical differentiation, and limited model expression ability [48]. To overcome these challenges, we introduce the HSM. The goal of the HSM is to map high-dimensional, noisy raw sensor time series data to a low-dimensional, smooth hidden state space, thereby effectively representing the degradation state of the equipment, and providing reliable input for subsequent physical rule learning and RUL prediction.

The HSM uses a self-attention-based neural network for implementation. As shown in Figure 2, assume that the input raw sample data is $X = [x_1, x_2, \ldots, x_T] \in R^{T \times d_x}$, where $T$ is the length of the time series, and $d_x$ is the dimension of the original input features.

First, the original input feature $x_t$ is mapped to an embedding space through a linear layer to obtain the embedding vector $e_t$:

$$e_t = W^{(e)} x_t + b^{(e)} \tag{10}$$

Where, $W^{(e)}$ and $b^{(e)}$ are the weight matrix and bias vector of the linear embedding layer, respectively.

For the embedding sequence $E = [e_1, e_2, \ldots, e_T]$, the model calculates self-attention weights in the feature dimension. For the embedded feature vector $e_j \in R^T$ of each feature dimension $j$, define the query $Q_j$, key $K_j$ and value $V_j$:

$$Q_j = W^{(q)} e_j, \qquad K_j = W^{(k)} e_j, \qquad V_j = W^{(v)} e_j \tag{11}$$

Where, $W^{(q)}, W^{(k)}, W^{(v)}$ are the weight matrices of the query, key, and value, respectively.

Next, calculate the dot product of the query vector and the key vectors of other feature dimensions for each feature dimension, and normalize to obtain the attention weights:

$$\alpha_{j,m} = \text{softmax}\left(\frac{Q_j^T k_m}{\sqrt{d_k}}\right) = \frac{\exp(Q_j^T K_m / \sqrt{d_k})}{\sum_{n=1}^{d_e} \exp(Q_j^T K_n / \sqrt{d_k})} \tag{12}$$

Use the attention weights to weight and average the value vectors to get the output $z_j \in R^T$ of the feature dimension $j$:

$$z_j = \sum_{m=1}^{d_e} \alpha_{j,m} V_m \tag{13}$$

The outputs of all feature dimensions are concatenated into $Z = [z_1, z_2, \ldots, z_{d_e}]$.

The output $Z = [z_1, z_2, \ldots, z_{d_e}]$ of the self-attention layer is subjected to a non-linear transformation through a feedforward neural network (FFN) to further enhance the expressive ability of the model. The FFN contains two linear layers and an activation function (ReLU):

$$h' = \text{FFN}(z_t) = W^{(f2)} \text{ReLu}(W^{(f1)} z_t + b^{(f1)}) + b^{(f2)} \tag{14}$$

Where, $W^{(f1)}, W^{(f2)}, b^{(f1)}, b^{(f2)}$ are the weight matrix and bias vector of the two linear layers in the FFN, respectively.

To alleviate the gradient vanishing problem and accelerate the training process, residual connections and layer normalization (LN) are added around the self-attention layer and the FFN layer. Finally, the output of the HSM is a hidden state vector sequence $h \in R^{d_h}$.

Through the above steps, the HSM converts the original, high-dimensional, noisy data $X$ into a smooth, continuous hidden state sequence $h$.

*3.2.2 RUL predictor*

RUL prediction is the ultimate goal of the model. After the HSM extracts the smooth continuous hidden state, the RUL predictor will fuse these hidden states with the corresponding time information and transforming them into accurate RUL estimates. The RUL predictor adopts a structural design that combines deep neural networks and an adaptive weighting mechanism.

The input of the RUL predictor is the concatenation of the hidden state vector $h$ output by the HSM and the corresponding equipment running time $t$, which can be expressed as:

$$c = [h, t] \tag{15}$$

Where, $[\cdot,\cdot]$ represents the vector concatenation operation.

The concatenated feature vector $c$ passes through a deep neural network (DNN) composed of four fully connected layers, with the Tanh used as the activation function between each layer:

$$z^{(1)} = tanh(W^{(p1)}c + b^{(p1)}) \tag{16}$$
$$z^{(2)} = tanh(W^{(p2)}z^{(1)} + b^{(p2)}) \tag{17}$$
$$z^{(3)} = tanh(W^{(p3)}z^{(2)} + b^{(p3)}) \tag{18}$$
$$o = W^{(p4)}z^{(3)} + b^{(p4)} \tag{19}$$

The output $o$ of the DNN is multiplied element-wise (Hadamard product) with the adaptive weighting coefficient $\rho$, to achieve adaptive weighting of each prediction component:

$$y = o \odot \rho \tag{20}$$

Where, $\odot$ represents the element-wise multiplication operation.

Finally, the RUL prediction value is obtained by summing the weighted components. We use $\hat{u}(h, t)$ to represent the RUL prediction value predicted by the model at the running time $t$, corresponding to the hidden state $h$:

$$\hat{u}(h, t) = \sum_{i=1}^{n} y_i \tag{21}$$

*3.2.3 Physics-Guided Regulator (PGR)*

The core goal of the PGR is to discover and describe the physical laws governing the change of RUL with the hidden state. Given the hidden state vector $h$ extracted by the hidden state mapper HSM and the current running time $t$, we take the RUL of the equipment as the state variable $u(h, t)$ related to the description of the equipment degradation, and model its evolution law with time and hidden state. From formula (1), it can be known that the dynamic evolution of the equipment degradation process can be expressed as the following PDE form:

$$\frac{\partial u(h,t)}{\partial t} - \mathcal{N}\left[u, \frac{\partial u}{\partial t}, \nabla_h u, \nabla_h^2 u, \dots, \nabla_h^k u\right] = 0, \quad h \in \Omega, t \in [0, T] \tag{22}$$

Where, $\nabla_h u$ represents the gradient of $u$ with respect to the hidden state $h$:

$$\nabla_h u = \left(\frac{\partial u}{\partial h_1}, \frac{\partial u}{\partial h_2}, \dots, \frac{\partial u}{\partial h_{d_h}}\right) \tag{23}$$

$\nabla_h^k u$ represents the $k$-th order spatial derivative of u with respect to $h$; $\mathcal{N}[\cdot]$ is a non-linear differential operator, describing the complex relationship between the state variable and its time and space derivatives; $\Omega$ represents the hidden state space.

Since the physical laws of the degradation process are usually difficult to express explicitly through analytical methods, we designed a PGR (represented by $\mathcal{P}_\theta$) so that it can adaptively learn

and approximate the non-linear operator $\mathcal{N}[\cdot]$, and transform the equation into:

$$\frac{\partial u(\boldsymbol{h},t)}{\partial t} - \mathcal{P}_{\boldsymbol{\theta}}(\hat{u}, \nabla_{\boldsymbol{h}}\hat{u}, \nabla_{\boldsymbol{h}}^2\hat{u}, \ldots, \nabla_{\boldsymbol{h}}^k\hat{u}) = 0, \quad \boldsymbol{h} \in \Omega, t \in [0,T] \tag{24}$$

Here $\mathcal{P}_{\boldsymbol{\theta}}$ is a parameterized neural network whose input contains the predicted state variable $\hat{u}$ and its spatial derivatives of various orders, and the output is the rate of change of the state variable with respect to time $\partial \hat{u}(\boldsymbol{h},t)/\partial t$.

To train the PGR, we first use Pytorch's automatic differentiation technology to calculate the derivatives of each order of the training samples, and construct a feature data set:

$$\mathcal{D} = \left\{\left(\hat{u}_i, \frac{\partial \hat{u}_i}{\partial t}, \nabla_{\boldsymbol{h}}\hat{u}_i, \nabla_{\boldsymbol{h}}^2\hat{u}_i, \ldots, \nabla_{\boldsymbol{h}}^k\hat{u}_i\right)\right\}_{i=1}^{M} \tag{25}$$

For each sample, the input feature vector of the PGR is:

$$(\hat{u}_i, \nabla_t \hat{u}_i, \nabla_{\boldsymbol{h}}\hat{u}_i, \nabla_{\boldsymbol{h}}^2\hat{u}_i, \ldots, \nabla_{\boldsymbol{h}}^k\hat{u}_i) \in R^d \tag{26}$$

Where $d$ is the total dimension of the input feature.

We use the MSE to define the physical loss between the PGR prediction output and the real-time derivative:

$$\mathcal{L}_{phy}(\boldsymbol{\theta}) = \frac{1}{M}\sum_{i=1}^{M}\left(\mathcal{P}_{\boldsymbol{\theta}}\left(\hat{u}_i, \frac{\partial \hat{u}_i}{\partial t}, \nabla_{\boldsymbol{h}}\hat{u}_i, \nabla_{\boldsymbol{h}}^2\hat{u}_i, \ldots, \nabla_{\boldsymbol{h}}^k\hat{u}_i\right) - \frac{\partial u_i}{\partial t}\right)^2 \tag{27}$$

Construct the data loss function:

$$\mathcal{L}_{data}(\boldsymbol{\theta}) = \frac{1}{M}\sum_{i=1}^{M}(u_i - \hat{u}_i)^2 \tag{28}$$

Get the total loss function of the final model:

$$\mathcal{L} = \mathcal{L}_{phy}(\boldsymbol{\theta}) + \mathcal{L}_{data}(\boldsymbol{\theta}) \tag{29}$$

After training, $\mathcal{P}_{\boldsymbol{\theta}^*}$ can approximately describe the dynamic evolution law of the equipment degradation process:

$$\frac{\partial u(\boldsymbol{h},t)}{\partial t} \approx \mathcal{P}_{\boldsymbol{\theta}^*}(\hat{u}_i, \nabla_t \hat{u}_i, \nabla_{\boldsymbol{h}}\hat{u}_i, \nabla_{\boldsymbol{h}}^2\hat{u}_i, \ldots, \nabla_{\boldsymbol{h}}^k\hat{u}_i) \tag{30}$$

3.3 Meta Parameters Update Framework

To achieve the goal that the proposed MKDPINN framework (whose all trainable parameters are represented by meta-parameters $\boldsymbol{\Phi}$ at the meta-learning level) can quickly adapt with only a small number of samples when facing new tasks, we designed a first-order optimization-based meta-learning framework. This framework aims to learn a set of optimized meta-parameters $\boldsymbol{\Phi}^*$, making them an efficient starting point for adapting to new RUL prediction tasks.

*3.3.1 Objective Function*

The ultimate goal of the proposed method is to find a set of meta-parameters $\boldsymbol{\Phi}^*$, so that starting from this point, the model can reach the lowest expected loss on the corresponding validation data $\mathcal{D}_p^{val}$ after performing $k$ steps of optimization on a small amount of training data $\mathcal{D}_p^{tr}$ of task $\mathcal{T}_p$ (parameters are represented as θ, using an inner-loop learning rate α and the Adam optimizer, starting from $\boldsymbol{\theta}_p^{(0)} = \boldsymbol{\Phi}$). Let $\text{Adapt}_k^{Adam}(\boldsymbol{\Phi}, \mathcal{D}_p^{tr}, \alpha)$ represent this $k$-step Adam adaptation process, which outputs the adapted parameters $\boldsymbol{\theta}_p^{(k)}$. From formula (6), the meta-optimization problem can be written as:

$$\boldsymbol{\Phi}^* = \arg\min_{\boldsymbol{\Phi}} E_{\mathcal{T}_p \sim P(\mathcal{T})}[F_p(\boldsymbol{\Phi})] \quad \text{where} \quad F_p(\boldsymbol{\Phi}) = \mathcal{L}_p(\text{Adapt}_k^{Adam}(\boldsymbol{\Phi}, \mathcal{D}_p^{tr}, \alpha); \mathcal{D}_p^{val}) \tag{31}$$

Where $\mathcal{L}_p$ is the total loss of our MKDPINN model on task $\mathcal{T}_p$ (data loss + physics loss, defined by equation (29)).

Directly optimizing equation (31) usually requires calculating the gradient propagated through the adaptation process. For MAML [29], this involves second-order derivatives (Hessian vector product), which is computationally expensive [35]. To circumvent this problem, we adopt an efficient first-order meta-learning algorithm, whose update mechanism evolves from Reptile [50]. This method does not directly calculate the gradient of $F_p(\boldsymbol{\Phi})$, but implicitly optimizes the objective through a clever update rule.

*3.3.2 First-Order Meta-Update Algorithm with Task Batch*

We adopt the following first-order meta-learning algorithm to update meta-parameters by batch processing tasks:

Meta-Training stage: Iterate through the following steps:

1. Task Batch Sampling: Randomly sample a mini-batch containing $B$ tasks from the source domain task distribution $P(\mathcal{T}): \{\mathcal{T}_p\}_{p=1}^B$.

2. Parallel Inner Loop Adaptation: For each task $\mathcal{T}_p (p = 1, \ldots, B)$ in the batch, independently perform the following operations:

   (a) Initialize task parameters: $\boldsymbol{\theta}_p^{(0)} = \boldsymbol{\Phi}$ (using the latest meta-parameters).

   (b) Use the Adam optimizer (with a learning rate of $\alpha$) to perform $k$ steps of updates on the training data $\mathcal{D}_p^{tr}$ of task $\mathcal{T}_p$ to minimize the loss $\mathcal{L}_p(\boldsymbol{\theta}; \mathcal{D}_p^{tr})$. Let $g_p^{(i)} = \nabla_{\boldsymbol{\theta}} \mathcal{L}_p(\boldsymbol{\theta}_p^{(i-1)}; \mathcal{D}_p^{tr})$ be the gradient of the $i$-th step with respect to the task parameter $\boldsymbol{\theta}$. The Adam update involves the exponential moving average estimation of the first-order moment $m$ and the second-order moment $v$ of the gradient:

$$m_p^{(i)} = \beta_1 m_p^{(i-1)} + (1 - \beta_1) g_p^{(i)}$$

$$v_p^{(i)} = \beta_2 v_p^{(i-1)} + (1 - \beta_2) \left(g_p^{(i)}\right)^2$$

$$\hat{m}_p^{(i)} = m_p^{(i)} / (1 - \beta_1^i)$$

$$\hat{v}_p^{(i)} = v_p^{(i)} / (1 - \beta_2^i)$$

$$\boldsymbol{\theta}_p^{(i)} = \boldsymbol{\theta}_p^{(i-1)} - \alpha \frac{\hat{m}_p^{(i)}}{\sqrt{\hat{v}_p^{(i)}} + \epsilon} \quad \text{for } i = 1, \ldots, k \tag{32}$$

Where $\beta_1, \beta_2 \in [0,1)$ are the exponential decay rates used to calculate the moment estimates, and $\epsilon$ is a small constant added for numerical stability. $\hat{m}_p^{(i)}$ and $\hat{v}_p^{(i)}$ are bias corrections for $m_p^{(i)}$ and $v_p^{(i)}$. Finally, the parameter $\boldsymbol{\theta}_p^{(k)}$ after adapting $k$ steps for task $p$ is obtained.

3. Outer Loop Meta-Parameter Update: Based on the adaptation results of all tasks in the batch, calculate the update amount of the meta-parameter $\boldsymbol{\Phi}$. The update rule is:

$$\boldsymbol{\Phi}^{new} = \boldsymbol{\Phi}^{old} + \eta \underbrace{\frac{1}{B} \sum_{p=1}^{B} \left(\boldsymbol{\theta}_p^{(k)} - \boldsymbol{\Phi}^{old}\right)}_{\text{Average update direction } \Delta\boldsymbol{\Phi}} \tag{33}$$

Where $\eta$ is the outer loop (meta) learning rate. This step moves the meta-parameter $\boldsymbol{\Phi}$ along the average vector direction from the initial point $\boldsymbol{\Phi}^{old}$ to the respective adapted parameters $\boldsymbol{\theta}_p^{(k)}$ of all tasks in the batch.

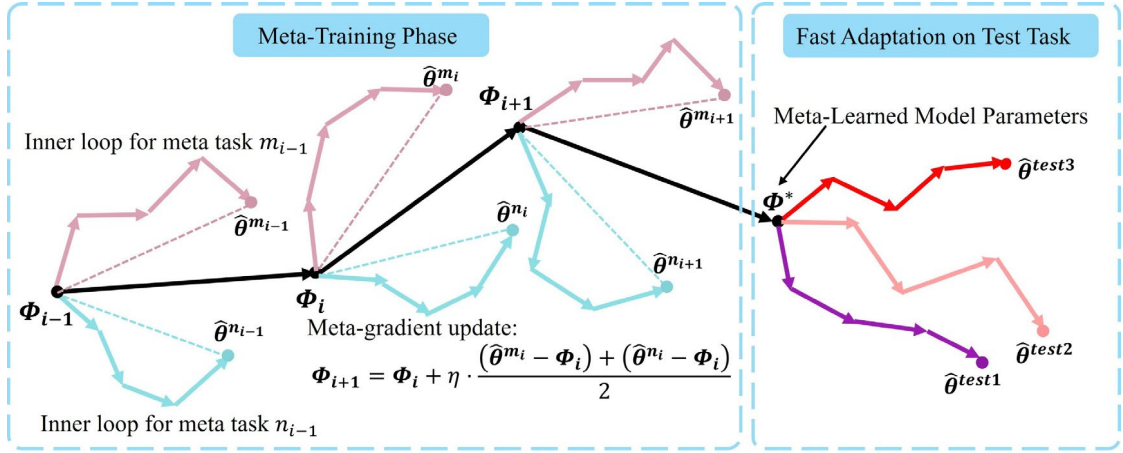

Fig. 3. Meta-Parameter Update of MKDPINN

Figure 3 intuitively depicts the update process of the first-order optimization-based meta-parameters $\boldsymbol{\Phi}$ in the proposed MKDPINN framework, which includes two core stages: meta-training and rapid adaptation. In the meta-training stage, the goal is to learn a set of optimized meta-parameters $\boldsymbol{\Phi}^*$ from a large number of source domain tasks, which possess good generalization ability and rapid adaptation potential. Specifically, in the $i$-th iteration, starting from the current meta-parameters $\boldsymbol{\Phi}_i$, the model samples a meta batch (Batch Size $B$, $B=2$ in the example figure) of meta-tasks ($m_i$ and $n_i$). For each task $p$ in the batch, inner loop optimization is performed: using $\boldsymbol{\Phi}_i$ as the initial parameters, using a small amount of training data $\mathcal{D}_p^{tr}$ from the task itself, the model parameters are updated independently through $k$ steps of gradient descent based on the Adam optimizer (as described in equation (32)), and finally, the parameters $\hat{\boldsymbol{\theta}}^{p(k)}$ adapted to the specific task are obtained (as shown by the endpoint $\hat{\boldsymbol{\theta}}^{m_i}$ and $\hat{\boldsymbol{\theta}}^{n_i}$ in the figure). After completing the inner loop adaptation of all meta tasks, the meta-parameters are updated in the outer loop: the vectors pointing from the current meta-parameters $\boldsymbol{\Phi}_i$ to the adapted parameters $\hat{\boldsymbol{\theta}}^{p(k)}$ of all tasks in the batch are calculated (i.e., the "task update vector" $\hat{\boldsymbol{\theta}}^{p(k)} - \boldsymbol{\Phi}_i$, as shown by the dashed lines in the figure), and the average of these vectors is calculated. The current meta-parameters $\boldsymbol{\Phi}_i$ are then updated along the direction of this average vector, with the step size controlled by the meta-learning rate $\eta$, so as to obtain the new meta-parameters $\boldsymbol{\Phi}_{i+1}$, whose update rule follows the equation $\boldsymbol{\Phi}_{i+1} = \boldsymbol{\Phi}_i + \eta \cdot (1/B) \sum_{p=1}^{B} (\hat{\boldsymbol{\theta}}^{p(k)} - \boldsymbol{\Phi}_i)$ (corresponding to equation (33)). By repeatedly iterating through this inner and outer loop process on many different meta task batches, the meta-parameters $\boldsymbol{\Phi}$ are driven to converge towards an "optimal initial point" $\boldsymbol{\Phi}^*$ that enables the model to perform well on various tasks with slight adjustments. Subsequently, when facing a new target domain task for the rapid adaptation stage, the final meta-parameters $\boldsymbol{\Phi}^*$ obtained from meta-training are used as the starting point for the new task model. Using the extremely small number of support samples $\mathcal{D}_{new}^{supp}$ provided by the new task, starting from $\boldsymbol{\Phi}^*$, perform $k'$ steps of gradient update as the meta-training inner loop, and the model parameters $\hat{\boldsymbol{\theta}}^{test}$ optimized for the characteristics of the new task can be quickly obtained for subsequent RUL prediction.

*3.3.3 Theoretical Analysis of the Meta-Update Direction*

The core of equation (33) lies in its update direction $\Delta \boldsymbol{\Phi} = 1/B \sum_{p=1}^{B} \left( \boldsymbol{\theta}_p^{(k)} - \boldsymbol{\Phi}^{old} \right)$.

Although the explicit meta-gradient $\nabla_{\Phi} F_p(\Phi)$ is not used, this direction implicitly contains the information needed to optimize the meta-objective (31). We can understand this by analyzing the expected properties of the update vector $d_p = \theta_p^{(k)} - \Phi^{old}$ for a single task.

Consider the $k$-step inner-loop optimization process. Even when using Adam, $\theta_p^{(k)}$ is obtained from $\Phi$, by iterating $k$ times depending on the gradient of the task $p$ loss function $\mathcal{L}_p$ with respect to the task parameter $\theta$ (calculated at the iteration point $\theta_p^{(i-1)}$). The vector $d_p$ represents the total displacement of the task parameters starting from $\Phi$ and moving $k$ steps under the loss-driven of task $p$.

We can draw on the idea of Taylor expansion to analyze the relationship between $d_p$ and the properties of $\mathcal{L}_p$ at the initial meta-parameter point $\Phi$. Let $g_{p,i} = \nabla_{\theta} \mathcal{L}_p\left(\theta_p^{(i-1)}; \mathcal{D}_{p,i}^{tr}\right)$ denote the (mini-batch) gradient used in the $i$-th step, $\bar{g}_{p,i} = \nabla_{\theta} \mathcal{L}_p(\Phi; \mathcal{D}_{p,i}^{tr})$ denote the gradient with respect to the task parameter $\theta$ at the initial meta-parameter point $\Phi$, and $\bar{H}_{p,i} = \nabla_{\theta}^2 \mathcal{L}_p(\Phi; \mathcal{D}_{p,i}^{tr})$ denote the corresponding Hessian matrix with respect to the task parameter $\theta$ (also evaluated at $\Phi$).

For simple SGD ($k$=1), $d_p = \theta_p^{(1)} - \Phi = -\alpha g_{p,1} \approx -\alpha \bar{g}_{p,1}$.

For SGD ($k$>1), from the formula (40) in reference [50], we can know that:

$$d_p = \sum_{i=1}^{k} \left(\theta_p^{(i)} - \theta_p^{(i-1)}\right) = -\alpha \sum_{i=1}^{k} g_{p,i} \tag{34}$$

Using the Taylor expansion $g_{p,i} \approx \bar{g}_{p,i} + \bar{H}_{p,i}\left(\theta_p^{(i-1)} - \Phi\right) \approx \bar{g}_{p,i} - \alpha \bar{H}_{p,i} \sum_{j=1}^{i-1} \bar{g}_{p,j} + O(\alpha^2)$, and substituting it into (34), we can obtain (ignoring the $O(\alpha^2)$ term):

$$d_p \approx -\alpha \sum_{i=1}^{k} \left(\bar{g}_{p,i} - \alpha \bar{H}_{p,i} \sum_{j=1}^{i-1} \bar{g}_{p,j}\right) = -\alpha \sum_{i=1}^{k} \bar{g}_{p,i} + \alpha^2 \sum_{i=1}^{k} \sum_{j=1}^{i-1} \bar{H}_{p,i} \bar{g}_{p,j} \tag{35}$$

Although the Adam update is more complex than SGD, making the precise Taylor expansion more difficult (because the update step size depends on the moment estimation of historical gradients), its update is essentially still an iteration process based on gradients. Therefore, it can be expected that $d_p = \theta_p^{(k)} - \Phi$ will still approximately capture a structure similar to equation (35), that is, it not only depends on the initial gradient $\bar{g}_{p,i}$ (evaluated at $\Phi$), but is also affected by the Hessian matrix $\bar{H}_{p,i}$ (evaluated at $\Phi$) and the interaction of previous gradients, especially when $k$>1.

Now consider the expectation of the outer loop update direction $E_{\mathcal{T}_p \sim P(\mathcal{T}), \{\mathcal{D}_{p,i}^{tr}\}}[d_p]$. Assume that each step $i$ of the inner loop uses different random mini-batch data $\mathcal{D}_{p,i}^{tr}$ from task $p$. Take the expectation of equation (35):

$$E[d_p] \approx -\alpha \sum_{i=1}^{k} E\left[\bar{g}_{p,i}\right] + \alpha^2 \sum_{i=1}^{k} \sum_{j=1}^{i-1} E\left[\bar{H}_{p,i} \bar{g}_{p,j}\right] \tag{36}$$

Define the Average Gradient $\text{AvgGrad}(\Phi) = E_{\mathcal{T}_p, \mathcal{D}_p^{tr}}[\nabla_{\theta} \mathcal{L}_p(\Phi; \mathcal{D}_p^{tr})] = E\left[\bar{g}_{p,i}\right]$ (assuming that the expected gradients in different steps are the same). This term drives $\Phi$ towards the minimization direction of the average loss of all tasks (evaluated at the point $\Phi$) (similar to joint training).

Define the Average Gradient Inner-product term $\text{AvgGradInner}(\Phi) = E_{\mathcal{T}_p, \mathcal{D}_{p,i}^{tr}, \mathcal{D}_{p,j}^{tr}}\left[\bar{H}_{p,i} \bar{g}_{p,j}\right]$ (where $i \neq j$, and $\mathcal{D}_{p,i}^{tr}, \mathcal{D}_{p,j}^{tr}$ come from the same task $\mathcal{T}_p$, $\bar{H}_{p,i}$ and $\bar{g}_{p,j}$ are both evaluated at

$\boldsymbol{\Phi}$). As analyzed in reference [50] (equations 29-32), this type of term is related to maximizing the (expected) inner product between gradients of different data batches within the task (evaluated at $\boldsymbol{\Phi}$), that is, $\nabla_{\boldsymbol{\Phi}} E_{\mathcal{T}_p, \mathcal{D}_{p,1}, \mathcal{D}_{p,2}} [\overline{g}_{p,1} \cdot \overline{g}_{p,2}]$ (note that the derivative here is with respect to $\boldsymbol{\Phi}$). Optimizing this term helps to improve the model's ability to quickly adapt from $\boldsymbol{\Phi}$ (in-task generalization).

Substitute these definitions into equation (36) to get:

$$E[d_p] \approx -(k\alpha)\text{AvgGrad}(\boldsymbol{\Phi}) + \left(\frac{k(k-1)}{2}\alpha^2\right)\text{AvgGradInner}(\boldsymbol{\Phi}) \tag{37}$$

Therefore, the average direction $\Delta\boldsymbol{\Phi} = (1/B)\sum_{p=1}^{B} d_p$ used in our outer loop update step (33), has an expectation of $E[\Delta\boldsymbol{\Phi}] = E[d_p]$. This means that our update step $\boldsymbol{\Phi}^{new} = \boldsymbol{\Phi}^{old} + \eta\Delta\boldsymbol{\Phi}$ is essentially performing an approximate gradient ascent (if $\eta > 0$). The key is that the update direction simultaneously contains components pointing towards $-\text{AvgGrad}$ (minimizing the average loss) and $+\text{AvgGradInner}$ (maximizing the in-task gradient alignment to promote rapid adaptation). Their relative weights are implicitly determined by $k$ and $\alpha$.

In this way, the proposed first-order meta-learning algorithm based on Adam and task batches, implicitly optimizes the meta-learning objective (31) without calculating second-order derivatives. The learned meta-parameter $\boldsymbol{\Phi}$ can balance the average performance on all tasks and the potential for rapid adaptation on new tasks.

## 4. Experiment

### 4.1 Case study 1: Real Industrial Scenarios

To validate the proposed RUL prediction methodology in a relevant industrial setting, experiments were conducted on critical assets within a large-scale iron ore processing system (an iron ore concentrator). The experimental objects are two 150ZJ-I-A65 type slurry pumps, which have a rated power of 178kW, a designed head of 63.2m, a rated flow of 470m³/h, and a working speed of 980rpm. Among them, pump No. 1 fully recorded two full life-cycle operating data including installation and commissioning, normal operation, degradation failure, and re-launch after maintenance, and pump No. 2 provided complete degradation process data from initial operation to failure. The on-site operation of the pump is shown in Figure 4. Vibration sensors are installed at the drive end, non-drive end, and the drive end and free end of the motor to collect vibration signals with a sampling frequency of 12.8kHz for each segment lasting 2.56 seconds, and periodic sampling is performed at intervals of 10 minutes.

Figure 5 shows the wear condition of the pump casing after the first life cycle of pump No. 1.

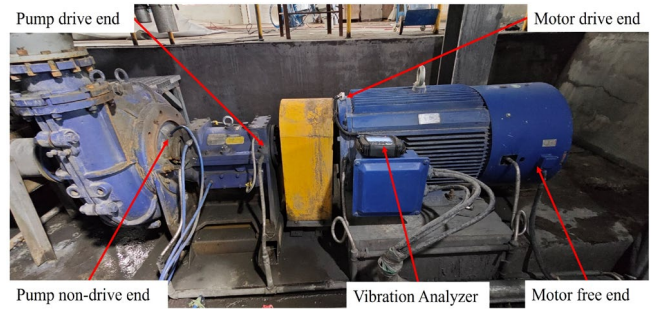

Fig. 4. Operational Status and Vibration Data Acquisition of Slurry Pump in Industrial Site

It can be seen that the pump casing has been worn through. In addition to the perforations, erosion-corrosion marks also appear in other areas on the inner wall of the pump casing. This shows the harsh operating environment of the slurry pump. Figure 6 shows the original vibration signal waveform of slurry pump No. 1 in two complete life cycles. Compared with the continuous and stable vibration signals usually obtained in a controlled laboratory environment, these data from real industrial scenarios show complexity. The non-regular idling (amplitude significantly reduced) and shutdown (amplitude close to zero) periods are clearly visible in the figure. Further analysis shows that the longer idling periods are mainly concentrated in the initial stage of each life cycle (approximately the first 110 hours for cycle 1 and approximately the first 180 hours for cycle 2). This phenomenon is related to the commissioning and trial operation process of the equipment in the early stage, or the need for the final-stage pump of the four-stage pumping system to wait for the upstream pump group to deliver the material. The intermittent shutdowns that occur in the subsequent operating stage are related to factors such as planned maintenance, production schedule adjustments, or upstream process interruptions. The existence of these non-steady-state conditions interferes with the continuous monitoring of the equipment degradation trend, thereby increasing the difficulty of the RUL prediction task.

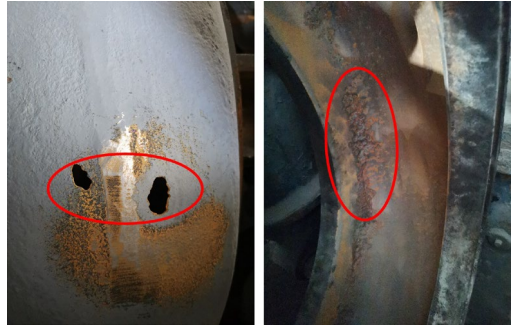

Fig. 5. Wear condition of the pump casing

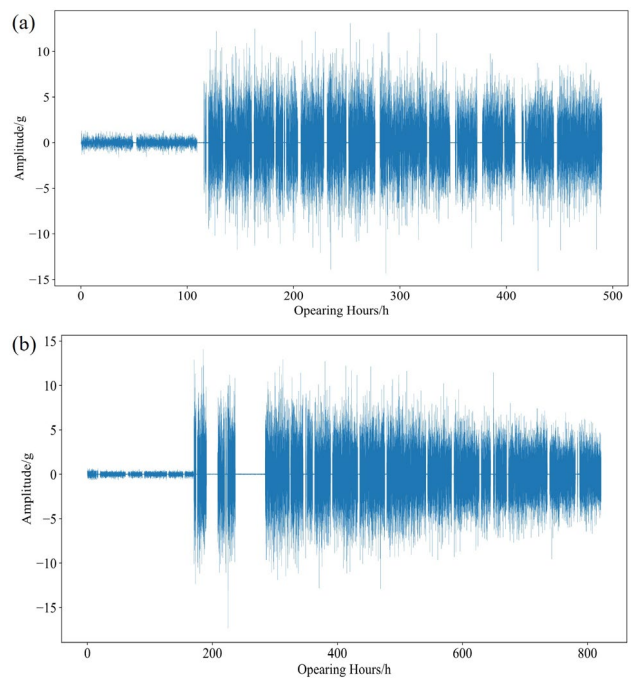

Fig. 6. Vibration Signals of Pump #1 during Two Life Cycles: (a) Cycle 1, (b) Cycle 2

*4.1.1 Data Preprocessing*

In this study, vibration signals are sampled by sensors at a frequency of 12800 Hz, and each sampling lasts for 2.56 seconds, generating a sequence with a length of 32768. To unify the time scale of analysis to the hour level, we extract features from the 6 continuous vibration signal sequences collected within each hour and calculate their average value as the feature representation for that hour.

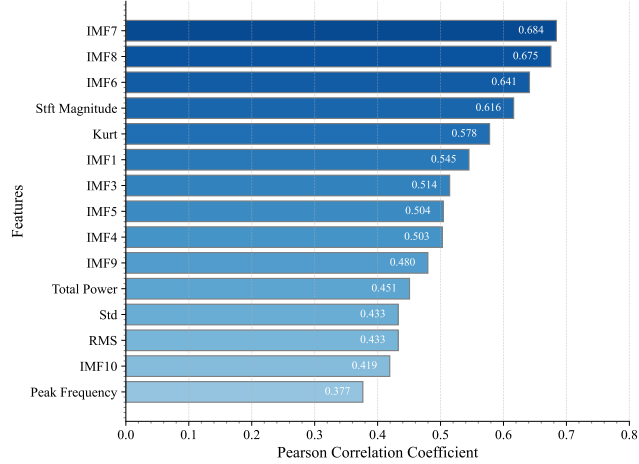

Fig. 7. Visualization of correlation analysis results

We extracted a wide range of candidate features from the original signal, includes mean, std, rms, max, min, peak-to-peak, Kurt, skewness, crest factor, shape factor, clearance factor, impulse factor, peak frequency, total power, spectral centroid, spectral kurt, spectral skew, STFT magnitude, and power of the $i$-th IMF component of the vibration signal. By calculating the Pearson correlation coefficient between each feature and the RUL, we screened out the 15 features with the strongest correlation. This screening process is based on the data analysis of the first complete life cycle of slurry pump No. 1 (the correlation ranking is shown in Figure 7).

To train the MKDPINN, we constructed training samples with a dimension of (15, 30). The first dimension of the sample represents the average value of the 15 selected features in the current hour, and the second dimension contains the historical sequence data of these features in the past 30 hours. In the allocation of sample labels, we use the actual remaining running hours of the equipment until failure as the RUL label. The label value of each sample represents the number of hours the equipment can effectively operate from the current moment to the failure moment. To accurately reflect the actual wear accumulation, we processed special working conditions: data during shutdown periods were identified and removed using the root mean square value of the signal (RMS < 0.1), and these shutdown times were not included in the calculation of RUL; for idling conditions (equipment running but without load), it is considered that there is no significant wear during the period, so its RUL label value remains the same as the previous effective running moment, that is, the RUL does not decrease with the passage of idling time.

*4.1.2 Model Parameter Configuration*

In this experiment, the model input dimension is set to 30×15, corresponding to the time step and the number of selected sensor features, respectively. We used two Attention-based Feature Extractor modules and applied a dropout rate of 0.1 to prevent overfitting. For the inner and outer loop optimization of meta-learning, the Adam optimizer is used for the inner loop, the learning rate (α) is set to 0.001, the number of update steps $k$ is 8, and the batch size is 64. The outer loop (meta-update) is based on a meta-batch ($B$) containing 5 meta-tasks, and the outer loop learning rate $\eta$ is set to 0.1.

The model is trained for a total of 50 epochs, of which 10% of the training data is divided into a validation set. During the training process, we save the optimal model weights according to the decrease in the validation set loss. The final model performance evaluation is performed in the 15-shot adaptation scenario.

To comprehensively evaluate the generalization ability of the model, we designed two test tasks:

1. Task 1 (Cross-Life Cycle RUL Prediction): The source domain data comes from the first life cycle of pump No. 1, and the target domain data is the second life cycle of the pump.

2. Task 2 (Cross-Machine Few-Shot RUL Prediction): This task represents a more challenging scenario and evaluates the model's RUL prediction performance for completely new equipment. The source domain data also comes from the first life cycle of pump No. 1, but the target domain data is taken from pump No. 2.

To comprehensively evaluate the performance of the proposed MKDPINN model and verify the effectiveness of its key components, we established three baseline models for comparative analysis. These models also form the basis of ablation experiments to measure the contribution of each module to the overall performance:

1. Base Learner: Serves as the basic model and adopts the same core network architecture and parameter configuration as MKDPINN, but does not integrate the PINN framework or the meta-learning strategy. Its performance represents the baseline level of relying solely on data-driven learning.

2. KDPINN: This model introduces the knowledge discovery-based PINN framework on the basis of the Base Learner. By comparing it with the Base Learner, the effect of PINN constraints on improving prediction accuracy can be evaluated.

3. Meta Learner: This model applies our proposed meta-learning framework on the Base Learner architecture, but omits the PINN part. Comparing it with the Base Learner and MKDPINN can reveal the contribution of the meta-learning strategy in improving the model's adaptability and generalization ability.

We use RMSE, MAE, $R^2$ as the evaluation criteria of the model performance, which are defined as follows:

$$RMSE = \sqrt{\frac{\sum_{i=1}^{n}(RUL_i - \widehat{RUL_i})^2}{n}} \tag{38}$$

$$R^2 = 1 - \frac{\sum_{i=1}^{n}(RUL_i - \widehat{RUL_i})^2}{\sum_{i=1}^{n}\left(RUL_i - \frac{1}{n}\sum_{i=1}^{n}RUL_i\right)^2} \tag{39}$$

$$MAE = \frac{1}{n}\sum_{i=1}^{n}|RUL_i - \widehat{RUL_i}| \tag{40}$$

*4.1.3 Experimental Results and Discussion*

Figure 8 shows the prediction results of each model in Task 1, and Table 1 shows the performance comparison of different models. The results indicate that: (a) The Base Learner model performed poorly, with considerable deviations between its predicted RUL and the actual RUL, especially with severe error accumulation in the later stages of equipment operation, and with substantial fluctuations in the prediction results. (b) Although the KDPINN model still had generally large prediction errors, thanks to the physical constraints introduced by the PINN framework, its predicted RUL showed a stronger monotonic decreasing trend compared to the Base Learner, demonstrating better prediction stability and better adherence to physical degradation laws, even if its performance was still insufficient. (c) The Meta Learner, by introducing the meta-learning framework, exhibited a clear advantage in adapting to target domain data, with a marked improvement in prediction accuracy compared to the previous two models and a large reduction in the magnitude of errors. (d) The proposed MKDPINN model combines the rapid adaptation capability of meta-learning and the physical constraints of PINN, and its prediction results are the most consistent with the actual RUL. Not only are the prediction results smooth and have the smallest errors, but they also maintain a good downward trend, demonstrating its superior performance and robustness in cross-life cycle RUL prediction tasks.

Figure 9 visually shows the distribution of key performance indicators (RMSE, MAE, $R^2$) for the four models after 10 independent repeated experiments through violin plots. It can be clearly observed from the figure that the proposed MKDPINN method has a considerable performance advantage in the cross-life cycle RUL prediction task compared to the baseline model and ablation models. Not only is the average prediction accuracy higher, but it also exhibits enhanced robustness

Table 1 Performance Comparison of Models on Task 1

| Model | RMSE | MAE | $R^2$ |
|---|---|---|---|
| Base Learner | 149.39 | 124.45 | 0.34 |
| KDPINN | 171.33 | 151.81 | 0.13 |
| Meta Learner | 43.60 | 34.27 | 0.94 |
| **MKDPINN** | **27.19** | **21.30** | **0.97** |

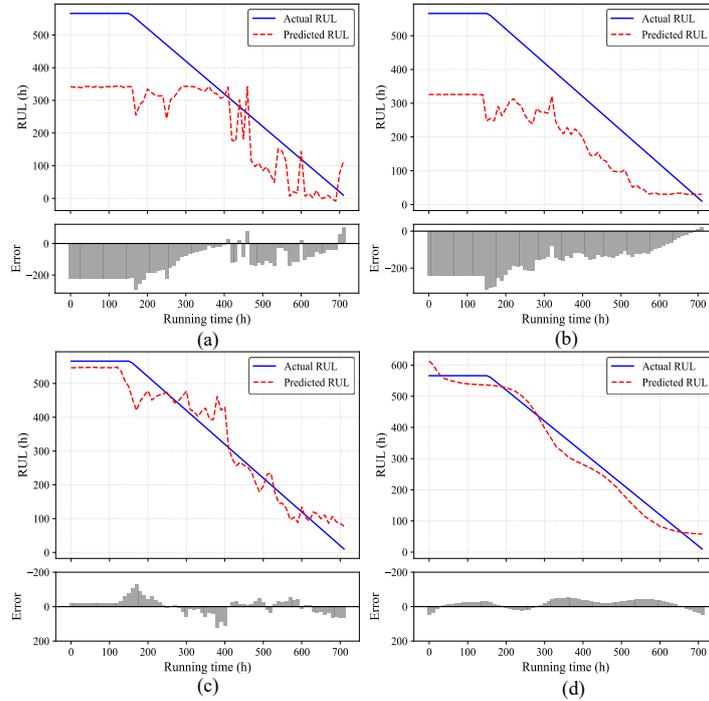

Fig. 8. Task 1 RUL Prediction Results: (a) Base Learner, (b)KDPINN, (c) Meta Learner, (d) MKDPINN.

and consistency in multiple experiments.

Figure 10 shows the wear condition of the pump casing of pump No. 2 used for Task 2. Its wear characteristics are manifested as local damage. A clear penetrating hole can be seen in the left image (outer wall perspective). The right image (inner wall perspective) reveals a larger range of surface damage inside the pump casing, showing typical erosion-abrasion features, including a large area of unevenness, pitting, and groove-like wear marks along the flow direction, reflecting the continuous erosion effect of the high-speed flow of the slurry on the material. Figure 11 presents the vibration signal waveform of pump No. 2 throughout its entire life cycle. Similar to the signal of pump No. 1 shown in Figure 6, the signal of pump No. 2 also clearly demonstrates the characteristics of discontinuous operation in a real industrial environment, including a low-amplitude idling period in the initial stage (approximately the first 110 hours) and high-amplitude vibrations in the subsequent working stage. However, compared to pump No. 1, the vibration signal of pump No. 2 shows more frequent and pronounced intermittent shutdown or idling periods after about 110 hours. This difference in operating mode, i.e., more frequent starts and stops or load fluctuations, indicates that pump No. 2 experienced different operational scheduling or process disturbances compared to pump No. 1 in actual working conditions, which also brings additional

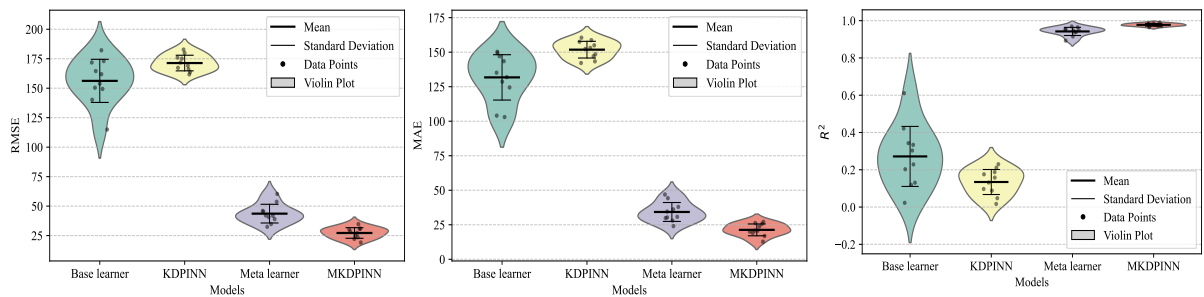

Fig. 9. Task 1: RUL Prediction Metric Comparison

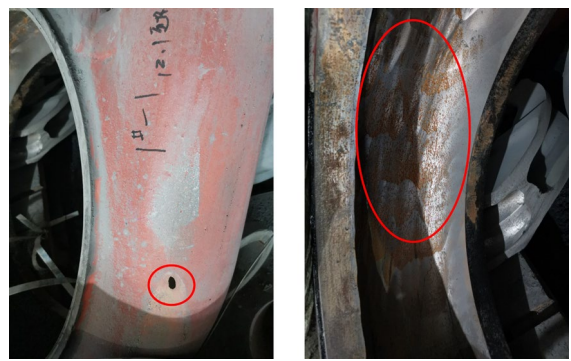

Fig. 10. Wear condition of Pump 2

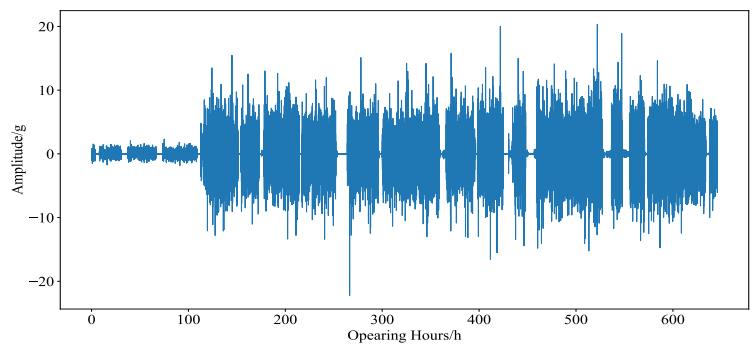

Fig. 11. Vibration Signals of Pump #2.

challenges to cross-machine RUL prediction.

Figure 12 presents the prediction results of the four models in Task 2; Table 2 shows the Performance Comparison of different models. It can be seen that the proposed MKDPINN model performs best in this more challenging cross-machine task. Its prediction results are closest to the actual RUL, and the prediction errors are the smallest and most stable. This highlights the important role of combining the rapid adaptation capability of meta-learning and the physical constraints of PINN in improving the RUL prediction accuracy and robustness of the model under few-shot, cross-machine conditions.

Table 2 Performance Comparison of Models on Task 2

| Model | RMSE | MAE | $R^2$ |
|---|---|---|---|
| Base Learner | 126.58 | 105.34 | 0.36 |
| KDPINN | 140.39 | 112.61 | 0.21 |
| Meta Learner | 44.11 | 33.88 | 0.91 |
| MKDPINN | **25.62** | **20.04** | **0.97** |

Figure 13 visually shows the statistical distribution of key performance indicators (RMSE, MAE, $R^2$) of each model after 10 independent experiments in Task 2 using violin plots. Comprehensively, the statistical results of Figure 13 strongly prove that even in cross-machine Few-Shot prediction tasks with data scarcity and obvious domain shift, the proposed MKDPINN method can still maintain high accuracy and high stability, superior to the baseline and ablation models.

Figure 14 explores the influence of the source of adaptation samples on the prediction performance of the MKDPINN model in the 15-shot cross-machine prediction scenario (Task 2), especially the performance of the model when it can only obtain health status information of the target equipment (pump No. 2). As shown in Figure 14(a), when the 15 adaptation samples are completely selected from the initial operation stage of the equipment and do not contain any clear degradation characteristics, the MKDPINN can still capture the basic trend of the RUL generally decreasing over time, rather than outputting invalid predictions. This prediction ability under zero-degradation sample adaptation highlights the intrinsic robustness of the model, which is mainly due to the general degradation prior knowledge and rapid adaptation ability given to the model by the PINN and meta-learning frameworks, and the early subtle state changes that the feature extractor may capture, which together enable the model to perform meaningful degradation trend extrapolation even when information is extremely scarce. Figure 15 shows the performance of MKDPINN for RUL prediction in 15-shot scenario using samples from different degradation ranges; Table 3 shows the performance metrics of MKDPINN for 15-shot RUL prediction with samples from different degradation stages. It can be seen that as the adaptation samples gradually include data from the later degradation stages, the prediction performance of the model is considerably improved, and the error is greatly reduced.

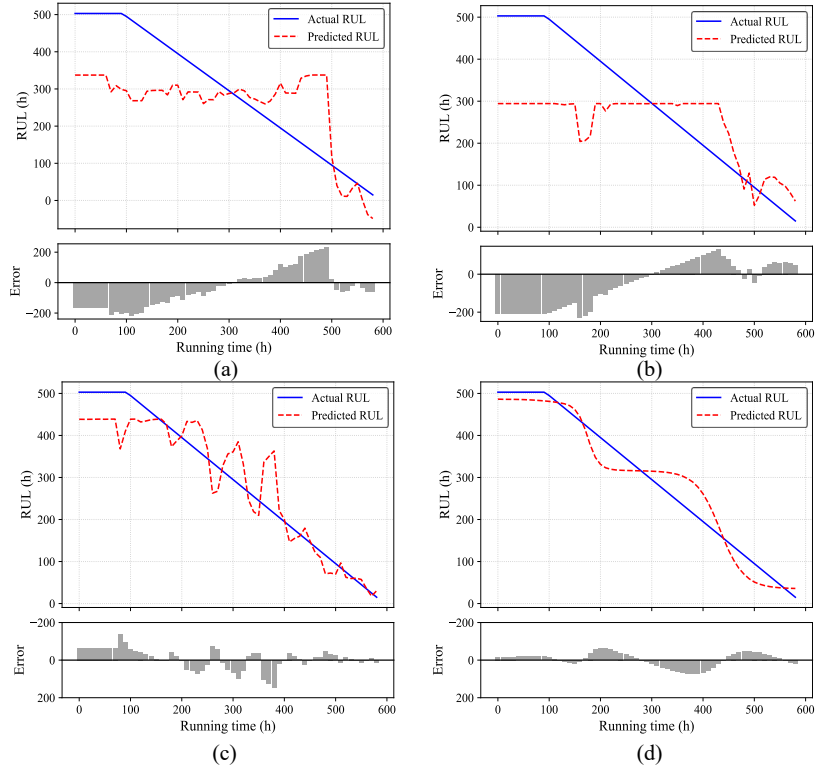

Fig. 12. Task 2 RUL Prediction Results: (a) Base Learner, (b) KDPINN, (c) Meta Learner, (d) MKDPINN.

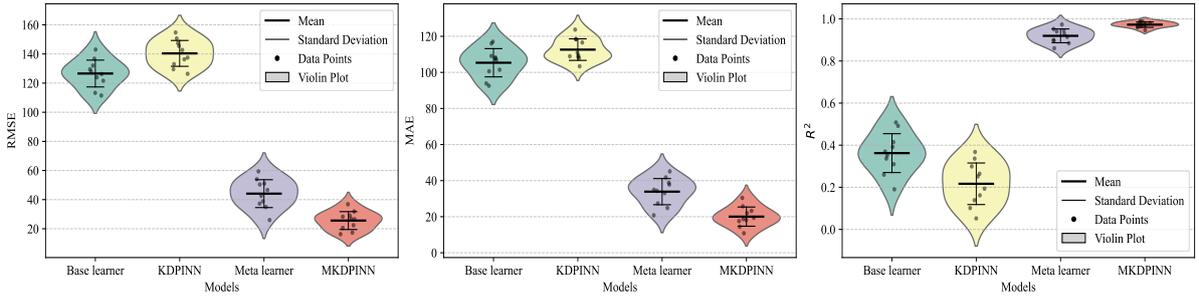

Fig. 13. Task 2: RUL Prediction Metric Comparison

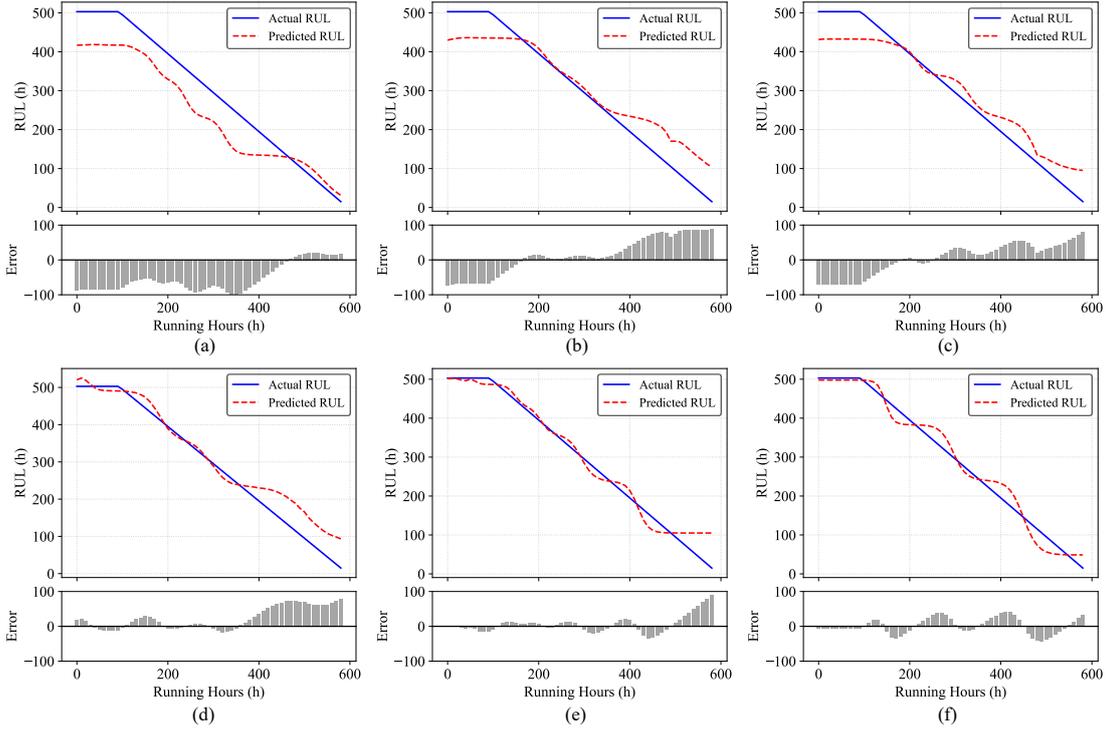

Fig. 14. RUL Prediction Results of MKDPINN in 15-shot Scenario Using Samples from Different Degradation Stages. (a) Using samples from normal operation (no degradation samples); (b) Using samples randomly selected from the first 20% of degradation samples; (c) Using samples randomly selected from the first 40% of degradation samples; (d) Using samples randomly selected from the first 60% of degradation samples; (e) Using samples randomly selected from the first 80% of degradation samples; (f) Using samples randomly selected from the entire (100%) degradation sample set.

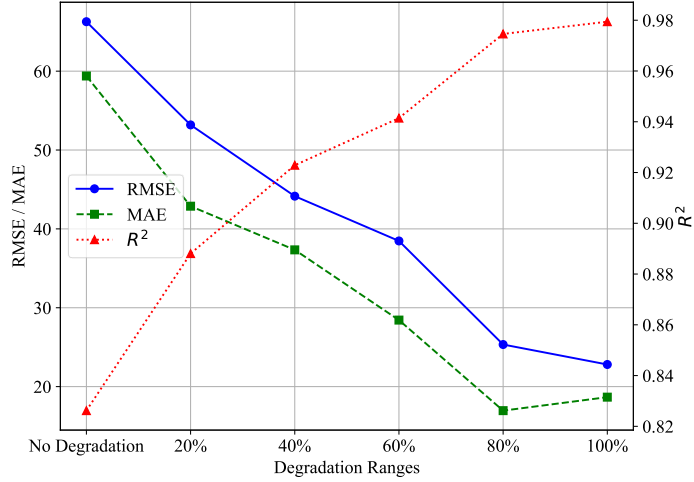

Fig. 15. Performance of MKDPINN for RUL Prediction in 15-shot Scenario Using Samples from Different Degradation Ranges

This finding proves the industrial application value of the MKDPINN model. For equipment that is newly put into operation or lacks historical failure data, the model can overcome the data accumulation period required by traditional methods and provide preliminary, trend-based RUL predictions based solely on initial healthy operating data, thereby supporting the early deployment of predictive maintenance strategies. This greatly reduces the reliance on complete "run-to-failure" datasets and reduces the cost and time constraints of data acquisition, especially suitable for scenarios where full life-cycle data cannot be obtained due to planned maintenance. More importantly, even if the accuracy of early predictions needs to be corrected by subsequent data, the RUL decline trend it provides itself constitutes an important early warning signal, which helps to pay attention to potential risks and adjust maintenance plans in a timely manner, reflecting the practicality and forward-looking nature of the method in data-limited real industrial environments. Although, as shown in Figures 14(b) to (f), incorporating degradation samples can considerably improve prediction accuracy, the basic prediction ability of the model under extreme information loss conditions has fully proven its superiority.

Table 3 Performance Metrics of MKDPINN for 15-Shot RUL Prediction with Samples from Different Degradation Stages

| Degradation Stages | RMSE | MAE | $R^2$ |
| --- | --- | --- | --- |
| No Degradation | 66.27 | 59.38 | 0.82 |
| First 20% of Degradation | 53.18 | 42.87 | 0.88 |
| First 40% of Degradation | 44.14 | 37.33 | 0.92 |
| First 60% of Degradation | 38.47 | 28.43 | 0.94 |
| First 80% of Degradation | 25.33 | 16.95 | 0.97 |
| 100% of Degradation | **22.80** | **18.66** | **0.97** |

We also evaluated the effect of the number of adaptation samples (i.e., "shot size") on the performance of the proposed MKDPINN model in the most challenging cross-machine RUL prediction task (Task 2). Figure 16 shows the RUL prediction results of MKDPINN when adapting the model using different shot sizes (5, 10, 15, 20 respectively). From Figures (a) to (d), it can be visually observed that as the number of samples used for adaptation increases, the predicted RUL becomes closer and closer to the actual RUL, and the magnitude of the prediction error also decreases, indicating that the prediction accuracy has a marked improvement.

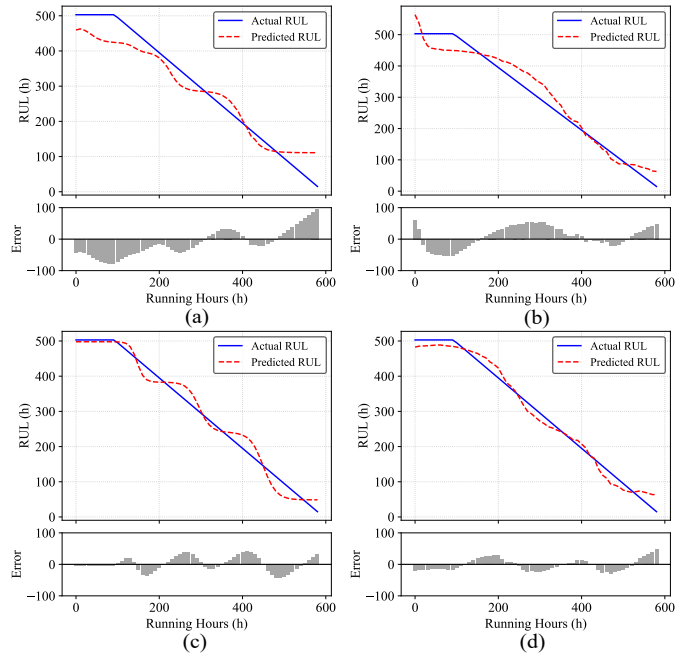

Fig. 16. RUL Prediction Results of MKDPINN for Cross-Machine RUL Prediction Tasks with Different Shot Sizes. (a) 5 shot; (b) 10 shot; (c) 15 shot; (d) 20 shot.

Figure 17 plots the curves of key performance indicators (RMSE, MAE, and $R^2$) versus shot size. This figure clearly reveals the quantified improvement in performance: as the shot size increases from 5 to 20, RMSE and MAE show a downward trend, while $R^2$ rises and approaches 1. This quantitatively confirms that increasing the number of adaptation samples can effectively improve the prediction accuracy and stability of the model.

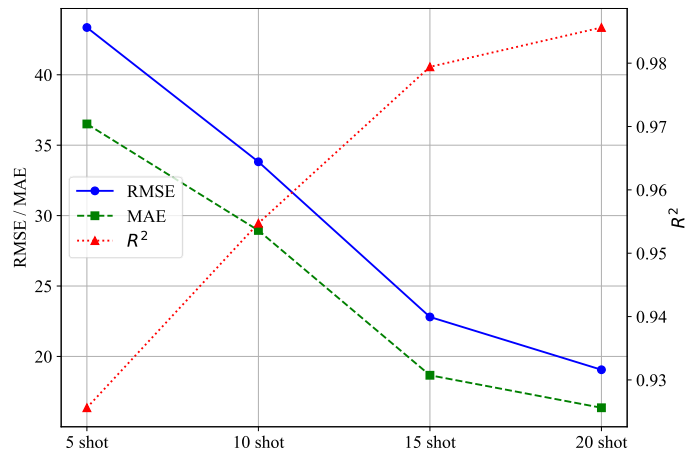

Fig. 17. Performance Metrics of MKDPINN for Cross-Machine RUL Prediction Tasks with Different Shot Sizes

Table 4 specifically lists the detailed performance indicator values under different shot sizes. The data shows that when the shot size is 5, the model's RMSE is 43.35, MAE is 36.50, and $R^2$ is

0.92. As the shot size increases, all indicators are continuously optimized. When the shot size reaches 20, the model achieves the best performance, RMSE drops to 19.05, MAE drops to 16.35, and $R^2$ is as high as 0.98. Combining the results of Figure 16, Figure 17, and Table 4, it is proven that only a small amount of labeled samples from the target equipment needs to be increased to considerably improve the accuracy and robustness of the MKDPINN model in the cross-machine Few-Shot RUL prediction task.

Table 4 RUL Prediction Performance of MKDPINN Under Different Test Scenarios

| Test Scenario | RMSE | MAE | $R^2$ |
|---|---|---|---|
| 5-Shot | 43.35 | 36.50 | 0.92 |
| 10-Shot | 33.81 | 28.94 | 0.95 |
| 15-Shot | 22.80 | 18.66 | 0.97 |
| 20-Shot | **19.05** | **16.35** | **0.98** |

4.2 Case Study 2: CMAPSS Dataset

*4.2.1 Dataset Description*

The Commercial Modular Aero-Propulsion System Simulation (C-MAPSS) dataset [51] is a benchmark dataset widely used in the PHM field. This dataset contains multiple sets of run-to-failure data of turbofan engines generated by C-MAPSS software simulation. Each set of data contains multivariate time series records of multiple engine units, including 21 sensor readings and 3 operating setting values, reflecting the performance degradation process of the engine under different flight conditions and fault modes. The C-MAPSS dataset is usually divided into four subsets (FD001, FD002, FD003, FD004). The details of this dataset are shown in Table 5, and the Detailed description of sensors in the C-MAPSS dataset is shown in Table 6.

Table 5 Description of the CMAPSS dataset

| Subset | Operational Condition | Fault Mode | Training Set Units | Test Set Units |
|---|---|---|---|---|
| FD001 | Single operating condition | Single fault mode | 100 | 100 |
| FD002 | 6 operating conditions | Single fault mode | 260 | 259 |
| FD003 | Single operating condition | 2 fault modes | 100 | 100 |
| FD004 | 6 operating conditions | 2 fault modes | 249 | 248 |

Table 6 Detailed description of sensors in the C-MAPSS dataset

| Number | Symbol | Description | Units |
|---|---|---|---|
| S1 | T2 | Temperature at the inlet of the fan | °R |
| S2 | T24 | Temperature at the LPC | °R |
| S3 | T30 | Temperature at the HPC | °R |
| S4 | T50 | Temperature at the LPT | °R |
| S5 | P2 | Fan inlet pressure | psia |
| S6 | P15 | Pressure within the bypass duct | psia |
| S7 | P30 | Pressure at the HPC outlet | psia |
| S8 | Nf | Fan's physical rotational speed | rpm |
| S9 | Nc | Core's physical rotational speed | rpm |
| S10 | epr | Engine pressure | – |
| S11 | Ps30 | Static pressure at HPC outlet | psia |
| S12 | phi | Fuel flow to Ps30 ratio | pps/psi |
| S13 | NRf | Corrected rotational speed of fan | rpm |
| S14 | NRc | Corrected rotational speed of core | rpm |
| S15 | BPR | Ratio of bypass air to core air | – |
| S16 | farB | Fuel-to-air ratio in the burner | – |
| S17 | htBleed | Enthalpy of the bleed air | – |
| S18 | Nf_dmd | Desired fan rotational speed | rpm |
| S19 | PCNfR_d | Desired corrected fan rotational speed | rpm |
| S20 | W31 | Coolant bleed from the HPT | lbm/s |
| S21 | W32 | Coolant bleed from the LPT | lbm/s |

*4.2.2 Data Preprocessing*

When constructing the samples for model training, we selected the following 14 sensor signals: ' S 2', ' S 3', ' S 4', ' S 7', ' S 8', ' S 9', ' S 11', ' S 12', ' S 13', ' S 14', ' S 15', ' S 17', ' S 20', ' S 21'. Considering that the C-MAPSS dataset contains a variety of different operating conditions, these changes in operating conditions can have a marked influence on sensor readings, which may mask the true equipment degradation trend. We adopted the Condition-Based Standardization (CS) method proposed in the literature [52] to process the selected sensor sequences.

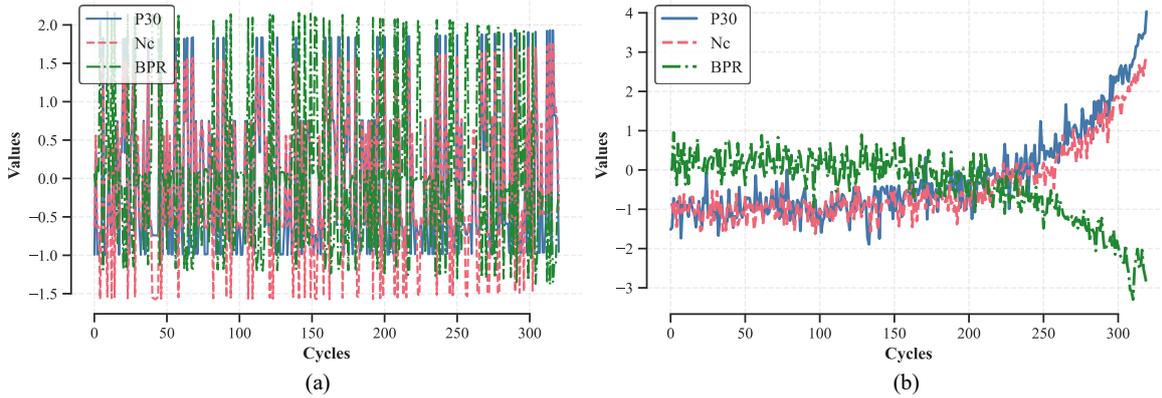

Fig. 18. Standardized Values of P30, Nc, and BPR Sensors: (a) Global Standardization, (b) Condition-Based Standardization

Different from the traditional Global Standardization (GS) method, the CS method first divides the data into different operating condition groups according to operating parameters, and then performs standardization calculations independently within each operating condition group. This strategy can effectively filter out signal fluctuations caused by operating condition switching, and more clearly reveal the intrinsic degradation patterns related to the equipment health status. Figure 18 visually compares the standardized values of three representative sensors (P30, Nc, BPR) after being processed by GS (Figure 18a) and CS (Figure 18b) methods. It can be clearly seen that the signal processed by CS (Figure 18b) presents smoother and more trend-oriented degradation characteristics, while in the signal processed by GS (Figure 18a), these degradation trends are largely submerged by dramatic operating condition fluctuations. Therefore, the use of the CS method helps to extract purer features that can better reflect the true degradation process.

After CS processing, we further applied the Exponentially Weighted Moving Average (EWMA) smoothing technique to each standardized sensor time series to further suppress high-frequency noise and random fluctuations in the signal, thereby more clearly highlighting the long-term degradation trend that reflects the slow changes in the equipment health status. EWMA achieves smoothing by assigning exponentially decreasing weights to historical data points. For a given sensor sequence $s = [s_1, s_2, \ldots, s_n]$, the formula for calculating EWMA is as follows:

$$s_t' = \rho * s_t + (1 - \rho) * s_{t-1}' \tag{41}$$

Where, $s_t'$ is the smoothed value at time point t, $s_t$ is the original sensor reading after CS at time point t, and $s_{t-1}'$ is the smoothed value at the previous time point. $\rho$ is the smoothing factor ($0 < \rho \leq 1$).

Finally, by applying the Sliding Window Method and setting the window length (i.e., the time step) to 15, a fixed-size window of 15 is used to slide along the time series data of each engine unit (including 14 selected sensors) with a stride of 1. At each sliding position, the 14 sensor readings on the 15 consecutive time points covered by the window are extracted to form a two-dimensional matrix sample with a shape of (15, 14). Where, the first dimension (15) represents the time step or sequence length, and the second dimension (14) represents the number of features (sensors) at each time step.

We followed the practice commonly used in the field to set the RUL labels. Specifically, we capped the actual RUL values: any RUL value exceeding 125 operating cycles was set to 125. This treatment helps the model focus on the degradation region of the equipment and maintains consistency with numerous comparative studies using C-MAPSS.

Regarding model evaluation, this experiment adopted a 0-shot testing scenario. This is because the C-MAPSS dataset itself provides clearly divided training and test sets, where the test set contains engine data that runs to an unknown time point and then terminates. According to the standard evaluation process of this dataset, we first train the MKDPINN model on the complete training set, and then directly apply the trained model to the test set for RUL prediction without using any samples from the test set (target domain) for any form of model fine-tuning or adaptation. Since the model has not been exposed to any target domain samples before evaluation, this process fully meets the definition of 0-shot learning. Choosing this standard 0-shot evaluation method also ensures that our results can be compared fairly and directly with a large number of existing studies on this dataset.

In addition to RMSE, this dataset also uses SCORE as an evaluation criterion, which is defined as follows:

$$SCORE = \begin{cases} \sum_{i=1}^{n} (e^{-\frac{\widehat{RUL}_i - RUL_i}{13}} - 1), & if\ \widehat{RUL}_i < RUL_i \\ \sum_{i=1}^{n} (e^{\frac{\widehat{RUL}_i - RUL_i}{10}} - 1), & if\ \widehat{RUL}_i \geq RUL_i \end{cases} \tag{42}$$

*4.2.3 Experimental Results and Analysis*

Figure 19 presents the overall results of the proposed MKDPINN model for RUL prediction on the four subsets (FD001, FD002, FD003, FD004) of the C-MAPSS dataset. This figure shows the comparison between the RUL prediction value and the corresponding actual RUL for all test engine units in each subset at the last time point of their available operating data. It can be seen from the figure that on all four subsets, the prediction values of the MKDPINN model show a high degree of consistency with the actual values. The prediction errors of most engines are small, and fluctuate around the zero value. This initially indicates that the model can achieve accurate prediction of the final RUL of the engine under different operating condition complexities and fault mode combinations, and has good generalization ability and robustness.

Figure 20 further shows the RUL prediction results of the MKDPINN model on each subset of

C-MAPSS. By randomly selecting a representative engine unit from each subset, the RUL prediction results for its entire operating cycle are plotted. Observation reveals that for these selected representative samples, the predicted RUL of the MKDPINN model can effectively present a downward trend. Especially after the engine enters the degradation stage, the predicted value and the true value are very close. Although there may be some fluctuations in the early stages or when operating conditions change, the model has successfully captured the key degradation dynamics on the whole, and can provide a relatively precise RUL estimate near the end of life, further verifying the model's adaptability on different datasets.

Table 7 compares the RUL prediction performance of the proposed MKDPINN model with a series of recently published advanced models on the four subsets of the C-MAPSS dataset. The results clearly show that the MKDPINN model exhibits excellent comprehensive performance. In terms of average performance, MKDPINN achieved the lowest average RMSE (12.71) and the lowest average SCORE (622.15) among all the compared models, which highlights its overall prediction accuracy and robustness under different complex working conditions and fault mode combinations. Among them, MKDPINN obtained the lowest RMSE values (13.85, 11.01, 13.60 respectively) on the three subsets of FD002, FD003, and FD004, and also obtained the lowest SCORE value (956.42) on the FD004 subset. This quantitative comparison result demonstrates the effectiveness and reliability of the MKDPINN model in RUL prediction tasks.

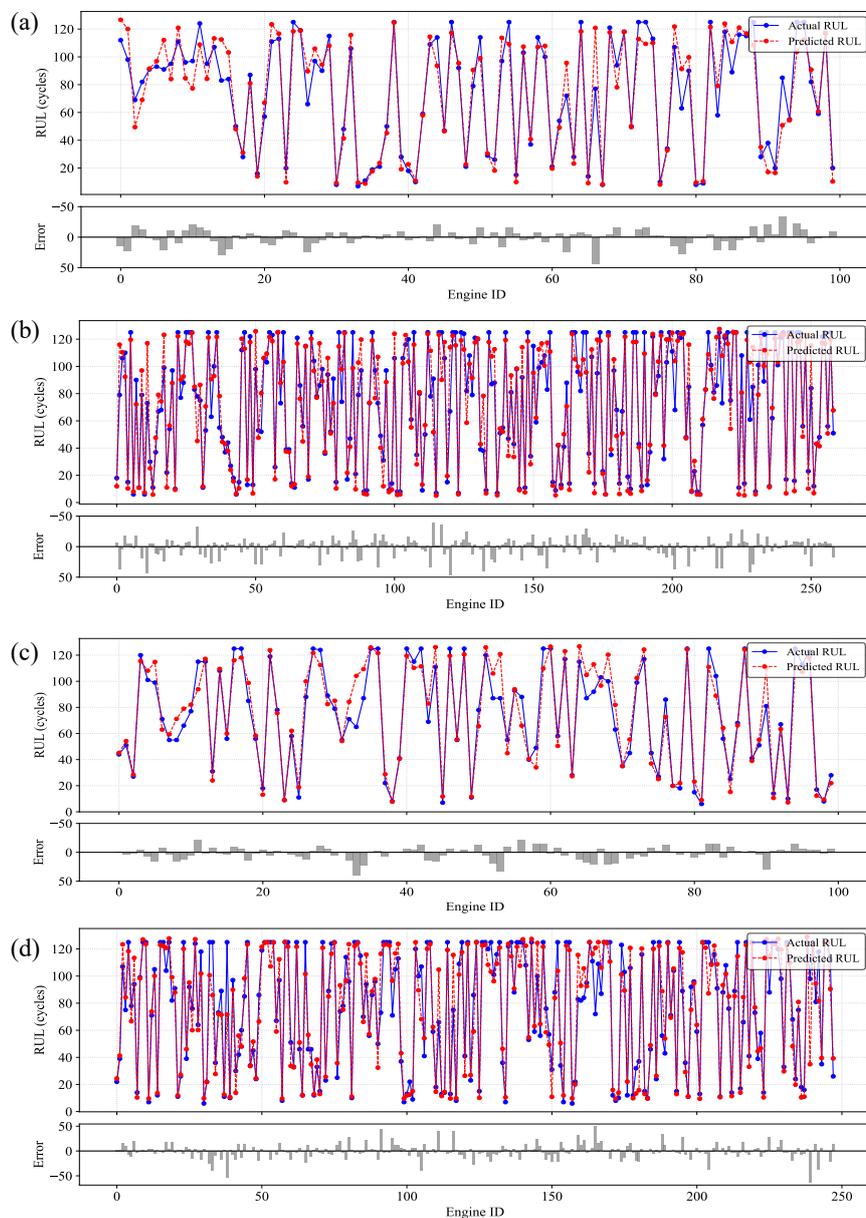

Fig. 19. MKDPINN RUL Prediction Results on C-MAPSS Dataset: (a) FD001, (b) FD002, (c) FD003, (d) FD004

Table 7 Performance Comparison of MKDPINN with Recently Published Models on C-MAPSS Dataset

| Model | FD001 | | FD002 | | FD003 | | FD004 | | Average | |
|---|---|---|---|---|---|---|---|---|---|---|
| | RMSE | SCORE | RMSE | SCORE | RMSE | SCORE | RMSE | SCORE | RMSE | SCORE |
| RVE[52] (2022) | 13.42 | 323.82 | 14.92 | 1379.17 | 12.51 | 256.36 | 16.37 | 1845.99 | 14.31 | 951.34 |
| ED-LSTM[53] (2023) | **9.14** | **53** | 18.17 | 1693 | 11.96 | 238 | 18.51 | 2160 | 14.45 | 1036.00 |
| AttnPINN[49] (2023) | 16.89 | 523 | 16.32 | 1479 | 17.75 | 1194 | 18.37 | 2059 | 17.33 | 1313.75 |
| MSTSDN[7] (2024) | 13.67 | 246.01 | 16.28 | 1342.59 | 13.66 | 258.92 | 17.33 | 1641.51 | 15.24 | 872.26 |
| ARR[54] (2024) | 11.36 | 192.22 | 18.97 | 2433.15 | 11.28 | **133.41** | 20.69 | 2842.44 | 15.58 | 1400.31 |
| Meta-Transformer[35] (2024) | 12.28 | / | 14.49 | / | 12.86 | / | 15.90 | / | 13.88 | / |
| MGCAL-UQ[55] (2025) | 11.63 | 180.61 | 17.16 | 1481.92 | 12.42 | 230.28 | 16.10 | 1278.56 | 14.33 | 792.84 |
| DAM[56] (2025) | 13.03 | 217 | 15.41 | **796** | 12.21 | 189 | 16.43 | 1029 | 14.27 | 557.75 |
| MKDPINN (proposed) | 12.36 | 273.19 | **13.85** | 1037.75 | **11.01** | 221.24 | **13.60** | **956.42** | **12.71** | **622.15** |

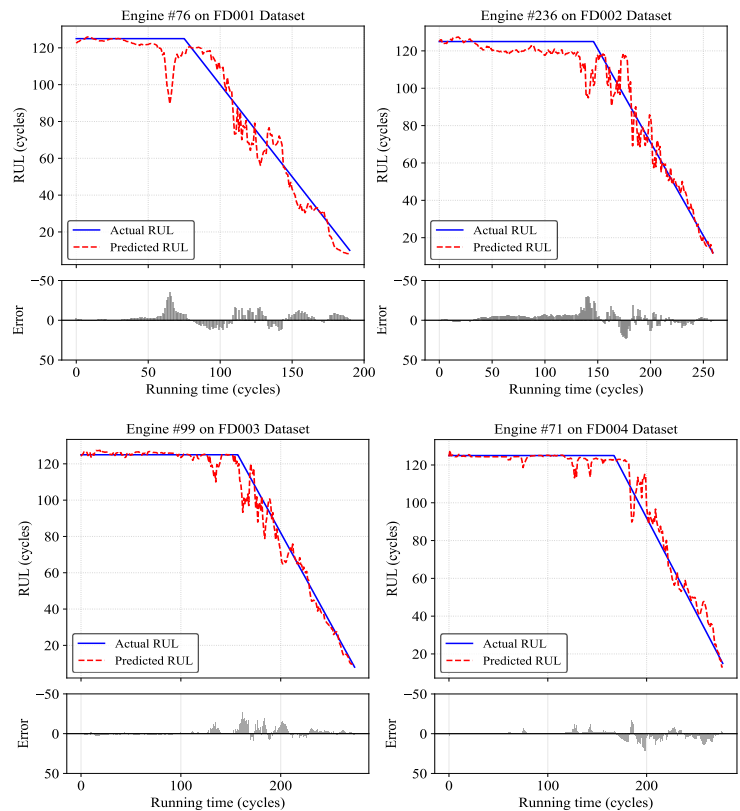

Fig. 20. Engine RUL Predictions on C-MAPSS Sub-datasets

## 5. Conclusion

We proposed MKDPINN to address two major challenges in practical industrial applications of rotating machinery RUL prediction in manufacturing systems: target domain sample scarcity and the difficulty of effectively incorporating physical laws due to a lack of explicit dynamic equations. We designed a knowledge discovery-based PINN framework, which learns the PDEs implicit in the equipment degradation process through the PGR and uses them to guide the training process of the model, improving the generalization ability of the model and its compliance with physical processes. At the same time, we embedded the knowledge discovery-based PINN framework into a first-order meta-learning strategy. Through learning on multiple meta-tasks, the meta-parameters of the model are aligned with the aggregated optimal parameters from each meta-task, enabling the model to quickly adapt to RUL prediction tasks under new equipment or new working conditions with extremely few target domain samples, substantially improving the model's practicality and deployment efficiency.

Through cross-life cycle and cross-machine few-shot RUL prediction experiments on a real industrial slurry pump dataset, and a comprehensive evaluation on the standard C-MAPSS benchmark dataset, the effectiveness of the MKDPINN method has been fully verified. The experimental results show that: 1. MKDPINN is substantially superior to baseline models that rely only on data-driven, only contain physics-informed information, or only use meta-learning in few-shot scenarios. It achieves the best performance on indicators such as RMSE, MAE, and $R^2$, demonstrating the synergistic advantage of combining physical knowledge discovery and meta-learning. 2. The method exhibits strong robustness when facing domain shift (such as cross-machine prediction). Even when only using the health status data of the target equipment for adaptation, it can provide meaningful RUL predictions, and the accuracy steadily improves as the adaptation samples cover more degradation stages or increase in number. 3. In the evaluation of the C-MAPSS dataset, MKDPINN achieved competitive results, exceeding the performance of existing advanced methods on multiple subsets, further confirming its applicability and accuracy.

In future research, we plan to explore how to extract more explicit and interpretable physical parameters or structures from the PDE learned by the PGR, and even combine prior domain knowledge to guide the PGR to learn specific forms of (partially known) physical equations, to achieve a deeper level of physics-informed information fusion.

## Acknowledgements

The work is supported by the National Natural Science Foundation of China (No. 51975100) and the Key Research and Development Program of Dalian (2024YF16PT026).

## Declarations

The authors declare that they have no financial interests or personal relationships that could have appeared to influence the work reported in this paper.